%% file: main.tex
\documentclass[11pt]{article}

\usepackage[utf8]{inputenc}
\usepackage[T1]{fontenc}
\usepackage[a4paper,margin=1in]{geometry}
\usepackage{amsmath}
\usepackage{amssymb}
\usepackage{bm}
\usepackage{graphicx}
\usepackage{float}
\usepackage{authblk}
\usepackage{natbib}
\usepackage{booktabs}
\usepackage[hidelinks]{hyperref}

\setcitestyle{authoryear,round,aysep={,},yysep={;}}

\title{Constitutive State-Space Modeling of Path-Dependent Plasticity:
A Resolution-Consistent and Parallelizable Computational Framework}

\author[1,2]{Rui Barreira}
\author[1]{Taylan Soydan}
\author[1,2]{Francesco Scipione}
\author[3]{Miguel A. Bessa}
\author[1]{Dirk Mohr\thanks{Corresponding author: dmohr@ethz.ch}}

\affil[1]{Chair of Artificial Intelligence in Mechanics and Manufacturing,
ETH Zurich, Switzerland}
\affil[2]{inspire AG, Zurich, Switzerland}
\affil[3]{School of Engineering, Brown University, Providence,
Rhode Island, USA}

\date{}

\begin{document}

\maketitle

\begin{abstract}
Data-driven constitutive models for path-dependent plasticity are commonly
formulated using nonlinear recurrent neural networks, whose sequential state
evolution limits parallel training and whose predictions may depend on the
discretization of the applied strain path. We introduce a Constitutive State
Space (CSS) model that reformulates structured state-space dynamics as an
incremental constitutive operator. The strain increment is decomposed into
magnitude and direction: the loading direction drives the latent state-space
system, while the increment magnitude enters the zero-order-hold
discretization of its continuous-time linear recurrence. This
mechanics-tailored construction guarantees stationarity under zero
increments, strongly reduces sensitivity to strain-path resolution, and
retains the parallel-scan structure of S5 for efficient training on long
constitutive histories. The CSS and Minimal State Cell (MSC) architectures
are independently optimized and compared for four multiaxial path-dependent
material models including isotropic $J_2$ plasticity, pressure-sensitive foam
plasticity, and combined isotropic-kinematic hardening. CSS matches or
exceeds the prediction accuracy of the MSC, including approximately one order
of magnitude lower validation losses for the plastically incompressible
materials. More importantly, CSS maintains low errors across large changes in
strain-path discretization, whereas the MSC error increases substantially
when evaluated at coarser resolutions than used for training. CSS also trains
substantially faster for long sequences and requires fewer strain-stress
pairs to attain comparable or better accuracy. Analysis of the learned state
further reveals latent structure consistent with the dimensionality of the
underlying physical constitutive models. These results establish
mechanics-tailored structured state-space dynamics as a computational
framework for efficient and discretization-robust data-driven constitutive
modeling.
\end{abstract}


\input{sections/01-introduction}
\input{sections/02-formulation}
\input{sections/03-data-generation}
\input{sections/04-results}

\input{sections/05-conclusions}
\input{sections/06-backmatter}

\nocite{abueidda2021,heidenreich2024a,heidenreich2024b}

\bibliography{references}

\end{document}

%% file: sections/01-introduction.tex
\section{Introduction}\label{sec:1}

Data-driven constitutive modeling has emerged as a computational alternative to
prescribing closed-form internal-variable equations for complex material behavior
\citep{fuhg2024}. Early neural-network-based approaches to constitutive modeling date
back to the seminal work of \citet{ghaboussi1991}, followed by
\citet{theocaris1993}, \citet{yagawa1996}, \citet{ghaboussi1998}, and
\citet{waszczyszyn2001}. With the increasing availability of computational resources
and large training datasets, interest in data-driven constitutive modeling has grown
substantially over the past decade, starting with basic feedforward neural-network
formulations for elastic and elastoplastic behavior \citep[e.g.][]{le2015,bessa2017,li2020}.
The constitutive operator is learned from strain--stress histories and can subsequently
serve as a surrogate for the local material response in numerical simulations. Deep
learning methods are particularly attractive because they can represent nonlinear,
history-dependent mappings without requiring an explicit analytical choice of hardening
variables or evolution equations. The computational challenge, however, is not limited
to reproducing stress histories accurately: a learned constitutive update should also
possess numerical properties compatible with incremental mechanics, including stable
state evolution, limited sensitivity to the chosen strain-path discretization, and
efficient processing of long loading histories.

Many neural-network-enhanced constitutive models preserve mathematical structure drawn
from physical understanding. For example, physics-informed neural networks (PINNs)
constrain the loss function to satisfy physical assumptions or constraints, often in the
form of partial differential equations. Focusing on path-dependent plasticity,
\citet{he2022} proposed an energy-based PINN that relies on internal variables and
stochastic training to accurately describe elastoplastic behavior. \citet{qian2024}
deployed a PINN model to replace expensive crystal-plasticity finite element simulations
of TSV-Cu structures. More recently, \citet{jadoon2025} modeled plastic anisotropy in
metals using input-convex neural networks (ICNNs). Thermodynamics-informed neural
networks \citep[e.g.,][]{su2024} may be regarded as a specialized class of PINNs in
which the free energy and/or dissipation potential is learned. Recent applications also
include crystal-plasticity modeling by \citet{amirian2025}, who trained a
thermodynamics-informed neural network on high-fidelity representative-volume-element
simulations and combined it with K-means clustering to group material points exhibiting
similar local mechanical states, as first proposed by \citet{liu2016}.

Fully data-driven models learn exclusively from data, without a priori assumptions about
the network formulation, thereby remaining applicable to general material behaviors and
loading conditions. For path-dependent plasticity, \citet{mozaffar2019} and
\citet{gorji2020} showed that recurrent neural networks can describe nonlinear,
history-dependent material behavior through combinations of activation functions and/or
gating mechanisms that control the evolution of internal state variables. Several other
works followed based on existing recurrent neural network (RNN) architectures built upon
gated recurrent units (GRUs)
\citep[e.g.,][]{wu2020,qu2021,tancogne-dejean2021,friemann2023} or long short-term
memory (LSTM) cells \citep[e.g.,][]{wang2018,frankel2019,shah2022}.

Ultimately, custom RNN cells tailored to constitutive modeling applications were
developed in subsequent years. \citet{bonatti2021} introduced the Minimal State Cell
architecture, removing discontinuous gating mechanisms from classical RNN cells, and
later enforced self-consistency, i.e., \emph{approximate} resolution invariance with
respect to changes in strain-increment magnitude \citep{bonatti2022b}. These models have
demonstrated good performance across a wide range of materials, loading paths, and
applications, from serving as surrogates for analytical constitutive models to
representing complex representative-volume-element models, such as crystal-plasticity
models \citep{bonatti2022a}. This architecture has since been used in different forms,
e.g., in combination with graph neural networks to predict history-dependent deformation
and microstructural evolution \citep{hu2024}, and was later expanded into the PolyLMSC
framework \citep{hu2025}. More recently, \citet{heidenreich2025} introduced the Extended
Minimal State Cell, an evolution of the MSC architecture that accounts for rate- and
temperature-dependent effects in material behavior. For a comprehensive review of the
development and application of data-driven constitutive models, the reader is referred
to \citet{fuhg2024} and \citet{dornheim2024}.

Although the state-of-the-art MSC has proven robust across materials, it retains
limitations that are important from a computational-mechanics perspective. First, its
nonlinear recurrent loop processes increments sequentially, so training wall-clock time
grows directly with sequence length and the recurrence cannot exploit parallel sequence
evaluation. Second, its self-consistency with respect to strain-increment refinement is
approximate and is recovered only for sufficiently small increments. This can become
relevant whenever the same loading path is represented at different resolutions, as
occurs naturally in adaptive incremental simulations. Third, the MSC encoder is embedded
inside the recurrent loop and tightly coupled to the state update. Once the
representational capacity of this encoder saturates, increasing the state dimension
alone provides diminishing improvements \citep{bonatti2021}. These observations motivate
a constitutive architecture in which the dependence on increment magnitude is treated
directly at the level of the state evolution rather than only through a learned
nonlinear gate.

Structured state-space models provide such a framework by representing sequential data
through continuous-time latent dynamical systems that are discretized for the increments
at hand. State-space architectures have recently become powerful sequence models across
deep learning \citep[e.g.,][]{gu2021,gu2022}, computer vision
\citep[e.g.,][]{nguyen2022,zubic2024,zubic2025}, language modeling
\citep{gu2023}, physics \citep{xu2025}, and finance \citep{shi2024}. In contrast to a
generic application of a sequence model, constitutive modeling offers additional
structure: each input represents a physical strain increment with a well-defined
magnitude and direction, and a change in increment size should not be interpreted merely
as a change in sequence index. We therefore build on the Simplified Structured State
Space Sequence (S5) model \citep{smith2023}, whose linear recurrent core can be evaluated
by parallel scans, and modify its input and discretization specifically for incremental
mechanics.

The central contribution of this work is a mechanics-tailored reformulation of structured
state-space dynamics as a constitutive update, termed the Constitutive State Space (CSS)
model. The novelty is not the use of an S5 block alone. Instead, the physical strain
increment is decomposed into magnitude and direction, the direction acts as the forcing
of the latent dynamical system, and the increment magnitude enters directly into the
zero-order-hold discretization of its continuous-time state evolution. This construction
links the recurrent update to the physical deformation increment rather than to an
arbitrary sequence step. At the same time, the linear recurrent core preserves parallel
prefix-scan evaluation during training. The computational contributions examined in this
study are therefore: (i) a mechanics-tailored structured state-space formulation for
multiaxial path-dependent constitutive updates; (ii) stationary and strongly
resolution-robust state evolution under changes in strain-increment size; (iii)
parallelizable sequence training for long constitutive histories; and (iv) a systematic
assessment of accuracy, computational scaling, data efficiency, model capacity, and
learned-state structure against the MSC. The study deliberately focuses on the
constitutive-update level so that these numerical properties can be isolated from
spatial discretization and global solver effects. Section~\ref{sec:2} formulates the two
constitutive architectures, Section~\ref{sec:3} describes the generation of controlled
multiaxial benchmark histories, Section~\ref{sec:4} presents the computational
assessment, and Section~\ref{sec:5} summarizes the conclusions.

%% file: sections/02-formulation.tex
\section{Computational formulation of data-driven constitutive models}\label{sec:2}

In the context of path-dependent elasto-plasticity, the strain-driven modeling
challenge consists of predicting the current stress-tensor $\bm{\sigma}[t]$ from
the strain path $\bm{\varepsilon}[0\ldots t]$, starting from the initial
stress-free configuration ($\bm{\varepsilon}[0]=\bm{0}$) to the current state of
deformation $\bm{\varepsilon}[t]$. Formally, we write the mapping
\begin{equation}\label{eq:1}
\bm{\varepsilon}[0\ldots t] \rightarrow \bm{\sigma}[t].
\end{equation}
This scenario also reflects the typical experimental setting, in which a strain
history is applied to a specimen while the resulting stress history is measured.

In a state-variable-based modeling approach, it is postulated that a set of state
variables, represented by a vector $\bm{\chi}[t]$, memorizes the salient features
of the strain history up to the time $t$, such that the current stress
$\bm{\sigma}[t]$ can be computed from $\bm{\chi}[t]$ alone:
\begin{equation}\label{eq:2}
\bm{\varepsilon}[0\ldots t] \rightarrow \bm{\chi}[t]
\end{equation}
\begin{equation}\label{eq:3}
\bm{\chi}[t] \rightarrow \bm{\sigma}[t]
\end{equation}
In an incremental setting, with the loading increment
$\bm{\Delta\varepsilon} = \bm{\varepsilon}^{(n)} - \bm{\varepsilon}^{(n-1)}$, the
problem is rewritten as
\begin{equation}\label{eq:4}
f: \{\bm{\chi}^{(n-1)}, \bm{\Delta\varepsilon}\}
\rightarrow \{\bm{\chi}^{(n)}, \bm{\sigma}^{(n)}\}
\end{equation}
assuming that the information contained in the pair
$\{\bm{\chi}^{(n-1)}, \bm{\Delta\varepsilon}\}$ is equivalent to knowing the full
loading path $\bm{\varepsilon}[0\ldots t]$. In the data-driven approach pursued
here, the operator $f$ is approximated by a trainable neural network. Two such
approximations are considered: the Minimal State Cell (MSC) and a new
mechanics-tailored state-space model, which we term the Constitutive State Space
(CSS) model. Both are described in detail below.

\subsection{Minimal State Cell (MSC) model}\label{sec:2.1}

\begin{figure}[H]
  \centering
  \includegraphics[width=0.72\linewidth]{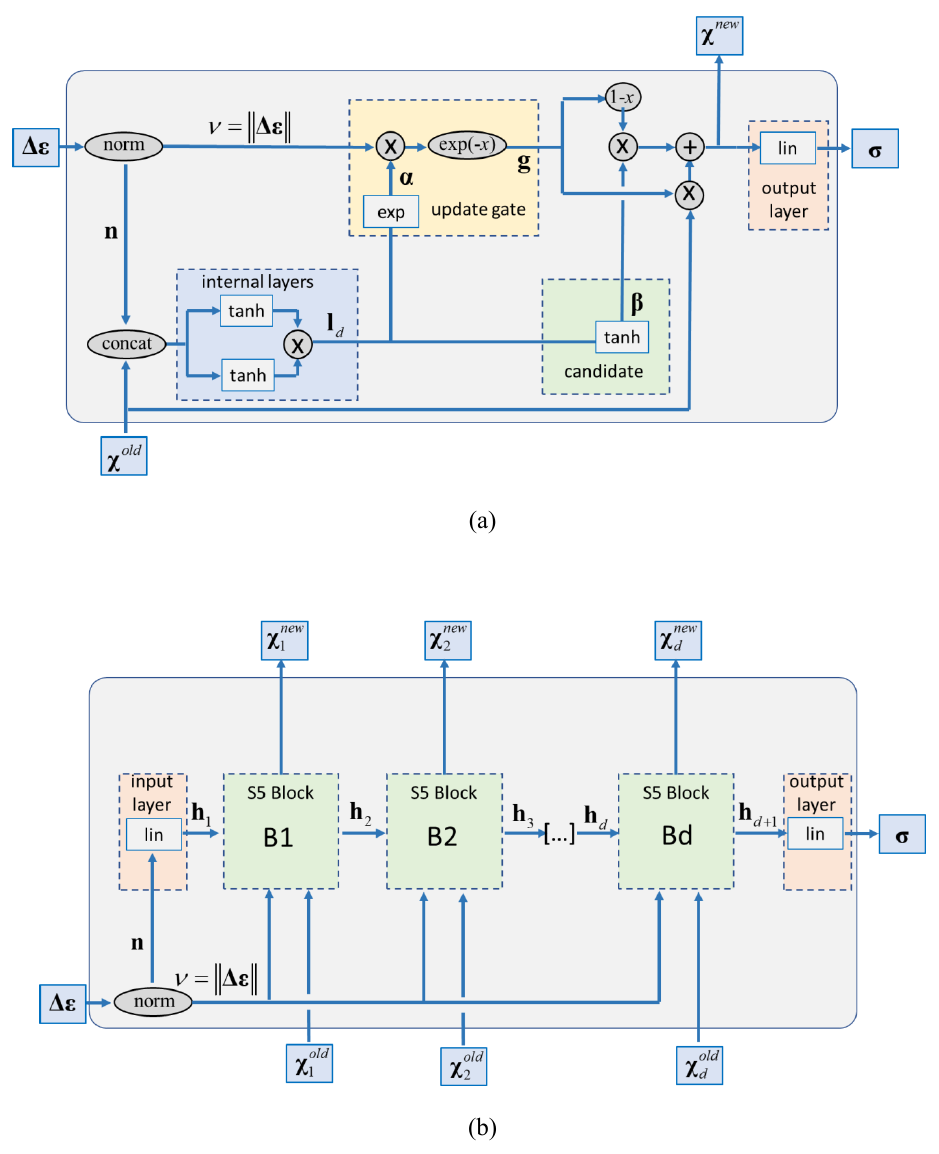}
  \caption{Recurrent neural network architectures: (a) Minimal State Cell (MSC) model featuring a gated architecture; (b) Constitutive State Space (CSS) model featuring $d$ stacked S5 blocks (B1, B2, \ldots, Bd).}
  \label{fig:1}
\end{figure}

The information flow of the MSC is illustrated schematically in
Fig.~\ref{fig:1}a. The operator $f$ is constructed by first decomposing the
strain input vector $\bm{\Delta\varepsilon} \in \mathbb{R}^{6}$ into its
magnitude ($L_2$-norm) $\nu = \|\bm{\Delta\varepsilon}\|$ and a unit vector
$\bm{n}$,
\begin{equation}\label{eq:5}
\bm{n} = \frac{\bm{\Delta\varepsilon}}{\|\bm{\Delta\varepsilon}\|}.
\end{equation}
The input to the first network layer is formed by concatenating the previous
state vector $\bm{\chi}^{(n-1)} \in \mathbb{R}^{s}$ (with $s$ the number of state
variables) with the current strain direction,
\begin{equation}\label{eq:6}
\bm{l}_0 = \bm{\chi}^{(n-1)} \oplus \bm{n},
\end{equation}
where $\oplus$ denotes vector concatenation. The vector
$\bm{l}_0 \in \mathbb{R}^{s+6}$ is then passed through $d$ fully connected
quadratic layers,
\begin{equation}\label{eq:7}
\bm{l}_i = \tanh\left(\bm{W}_i^a \bm{l}_{i-1} + \bm{b}_i^a\right) \odot
\tanh\left(\bm{W}_i^b \bm{l}_{i-1} + \bm{b}_i^b\right), \qquad i = 1,2,\ldots,d,
\end{equation}
with $\odot$ denoting the element-wise Hadamard product,
$\bm{W}_1^a, \bm{W}_1^b \in \mathbb{R}^{w \times (s+6)}$,
$\{\bm{W}_i^a, \bm{W}_i^b\} \in \mathbb{R}^{w \times w}$ for $i \geq 2$, and
$\{\bm{b}_i^a, \bm{b}_i^b\} \in \mathbb{R}^{w}$. From the output
$\bm{l}_d \in \mathbb{R}^{w}$, two vectors $\bm{\alpha} \in \mathbb{R}^{s}$ and
$\bm{\beta} \in \mathbb{R}^{s}$ are obtained using fully connected layers with
exponential and hyperbolic tangent activation functions, respectively,
\begin{equation}\label{eq:8}
\bm{\alpha} = \exp\left(\bm{W}_\alpha \bm{l}_d + \bm{b}_\alpha\right),
\end{equation}
\begin{equation}\label{eq:9}
\bm{\beta} = \tanh\left(\bm{W}_\beta \bm{l}_d + \bm{b}_\beta\right).
\end{equation}
The vector $\bm{\beta}$ acts as a candidate for the updated state vector, while
$\bm{\alpha}$ feeds into a gate $\bm{g} \in \left]0,1\right]$ that modulates the
state update as a function of the strain increment magnitude $\nu$,
\begin{equation}\label{eq:10}
\bm{g} = \exp\left[-\bm{\alpha}\nu\right]
\end{equation}
and
\begin{equation}\label{eq:11}
\bm{\chi}^{(n)} = \bm{g} \odot \bm{\chi}^{(n-1)}
+ (1 - \bm{g}) \odot \bm{\beta},
\end{equation}
with the new state $\bm{\chi}^{(n)} \in \mathbb{R}^{s}$. Finally, the stress is
computed via a linear output layer,
\begin{equation}\label{eq:12}
\bm{\sigma} = \bm{W}_\sigma \bm{\chi}^{(n)} + \bm{b}_\sigma,
\end{equation}
where $\bm{W}_\sigma$ and $\bm{b}_\sigma$ are the output weights and biases,
respectively. The architecture of the MSC model is fully determined by (i) the
number of state variables, $s$; (ii) the depth $d$, and (iii) the width $w$ of
its internal layers.

\subsection{Constitutive State Space (CSS) model}\label{sec:2.2}

The information flow in the CSS model is depicted in Fig.~\ref{fig:1}b. Like the
MSC, the CSS decomposes the input strain increment $\bm{\Delta\varepsilon}$ into
its magnitude $\nu$ and direction $\bm{n}$. It is routed through an input gate,
with
\begin{equation}\label{eq:13}
\bm{\alpha} = \exp\left(-\nu/\bm{\tau}\right)
\end{equation}
where $\bm{\tau} \in \mathbb{R}^{6}_{>0}$ is a vector of trainable parameters,
yielding the held direction $\tilde{\bm{n}} \in \mathbb{R}^{6}$,
\begin{equation}\label{eq:14}
\tilde{\bm{n}}^{(\mathrm{n})} = \bm{\alpha} \odot
\tilde{\bm{n}}^{(\mathrm{n}-1)} + (1 - \bm{\alpha}) \odot \bm{n}.
\end{equation}
The held direction is then encoded into a latent input vector via an affine
transformation,
\begin{equation}\label{eq:15}
\bm{h}_{1} = \bm{W}_{in}\tilde{\bm{n}} + \bm{b}_{in}.
\end{equation}
with $\bm{h}_{1} \in \mathbb{R}^{h}$, $\bm{W}_{in} \in \mathbb{R}^{h\times 6}$,
and $\bm{b}_{in} \in \mathbb{R}^{h}$ and the latent-space dimension $h$. The
vector $\bm{h}_{1}$ is then passed through a stack of $d$ Simplified Structured
State Space Sequence (S5) blocks.

Each S5 block $i$ ($i = 1,\dots,d$) consists of a linear recurrent core and a
nonlinear decoder. In the core, the components of the input vector
$\bm{h}_{\mathrm{i}}$ are first normalized to zero mean and unit variance,
\begin{equation}\label{eq:16}
\tilde{\bm{h}}_{\mathrm{i}} = \mathrm{LayerNorm}(\bm{h}_{i}).
\end{equation}
The evolution of the block's complex-valued state $\bm{\chi}_{i} \in
\mathbb{C}^{p}$ (where $p$ denotes the state dimension) is governed by an affine
ordinary differential equation. Applying a zero-order-hold discretization yields
\begin{equation}\label{eq:17}
\bm{\chi}_{i}^{(n)} = \tilde{\bm{A}}_{i}\,\bm{\chi}_{i}^{(n-1)}
+ \tilde{\bm{B}}_{i}\tilde{\bm{h}}_{\mathrm{i}},
\end{equation}
where $\tilde{A}_{i}$ and $\tilde{B}_{i}$ depend on the strain increment
magnitude $\nu$ (model input),
\begin{equation}\label{eq:16bis}
\tilde{\bm{A}}_{i} = \exp\left(\bm{A}_{i}\,\Delta_{\mathrm{i}}\right),
\end{equation}
\begin{equation}\label{eq:17bis}
\tilde{\bm{B}}_{i} = \bm{A}_{i}^{-1}\left(\tilde{\bm{A}}_{i}
- \bm{1}\right)\bm{B}_{i},
\end{equation}
Here, $\bm{A}_{i} \in \mathbb{C}^{p\times p}$ and $\bm{B}_{i} \in
\mathbb{C}^{p\times h}$ are trainable complex-valued matrices, with $\bm{A}_{i}$
being diagonal. The non-negative state-dependent time step vector
$\Delta_{\mathrm{i}}$ is defined as,
\begin{equation}\label{eq:18}
\Delta_{\mathrm{i}} = \nu \exp\left(w_{i}\right) \qquad \text{with} \quad
w_{i} \in \mathbb{R}^{p}.
\end{equation}
A first latent output of the state space block is then computed as,
\begin{equation}\label{eq:19}
\bm{y}_{i} = \mathrm{Re}(\bm{C}_{i}\,\bm{\chi}_{i}^{(n)})
+ \bm{D}_{i}\,\tilde{\bm{h}}_{\mathrm{i}}\quad ,
\end{equation}
with $\mathrm{Re}(\,\cdot\,)$ denoting the real part, $\bm{C}_{i} \in
\mathbb{C}^{h\times p}$ is the output update matrix, and $\bm{D}_{i} \in
\mathbb{R}^{h\times h}$ is a diagonal feed-through matrix.

The latent output is subsequently routed through a nonlinear decoder, which
applies a nonlinear activation (using a sigmoid linear unit), a residual
connection, and layer normalization:
\begin{equation}\label{eq:20}
\bar{\bm{h}}_{i} = \mathrm{SiLU}(\bm{y}_{i}) + \tilde{\bm{h}}_{\mathrm{i}},
\end{equation}
\begin{equation}\label{eq:21}
\hat{\bm{h}}_{i} = \mathrm{LayerNorm}\big(\bar{\bm{h}}_{i}\big).
\end{equation}
\begin{equation}\label{eq:22}
\bm{g}_{i} = \mathrm{SiLU}\big(\mathrm{W}_{\mathrm{enc,i}}
\left(\hat{\bm{h}}_{i}\right)\big).
\end{equation}
The final output of the $i^{\mathrm{th}}$ S5 block is,
\begin{equation}\label{eq:23}
\bm{h}_{i+1} = \bm{W}_{\mathrm{dec,\,i}}\,\bm{g}_{i} + \hat{\bm{h}}_{i},
\end{equation}
with $\bm{h}_{i+1} \in \mathbb{R}^{h}$, which also defines the input to the
subsequent S5 block. After the final block, the output $\bm{h}_{d+1}$ is passed
through a linear layer to produce the updated stress,
\begin{equation}\label{eq:24}
\bm{\sigma} = \bm{W}_{out}\bm{h}_{d+1}.
\end{equation}
The memory of the CSS is thus given by the state vectors of all blocks,
$\{\bm{\chi}_{1},\dots,\bm{\chi}_{d}\}$, each composed of $p$-dimensional real
and imaginary parts. The total count of trainable parameters includes the
components of the matrices $\{\bm{W}_{in}, \bm{b}_{in}, W_{\mathrm{enc},i},
\bm{A}_{i}, \bm{B}_{i}, \bm{C}_{i}, \bm{D}_{i}, W_{\mathrm{dec},i},
\bm{W}_{out}\}$, the learnable log-step $\Delta_{\mathrm{i}}$, input-hold time
constants $\tau$, and the two LayerNorm scale/bias pairs per block.

The architecture of the CSS is fully determined by (i) the latent input
dimension $h$, (ii) the number $d$ of S5-blocks, and (iii) the complex state
dimension $p$ of each block. Since each complex-valued component corresponds to
two real-valued state variables, i.e. the real and imaginary parts, the CSS
contains a total of $s = 2pd$ state variables.

The mechanics-specific feature of the CSS is the way in which increment size
enters this recurrence. For a fixed loading direction, the zero-order-hold
update is the exact discrete evolution of the linear state-space core over the
prescribed strain-increment magnitude. Subdividing such an increment therefore
changes the numerical resolution of the same latent evolution rather than
introducing an unrelated sequence step. Nonlinear decoding between stacked
blocks means that resolution robustness of the complete network remains a
property to be assessed numerically; Section~\ref{sec:4.3} therefore tests it
directly over changes in path resolution spanning almost two orders of
magnitude. This distinction is central to the proposed constitutive formulation
and separates it from applying an unmodified generic S5 sequence model to
strain-stress data.

%% file: sections/03-data-generation.tex
\section{Training data generation}\label{sec:3}

\subsection{Materials}\label{sec:3.1}

\begin{figure}[H]
  \centering
  \includegraphics[width=\linewidth]{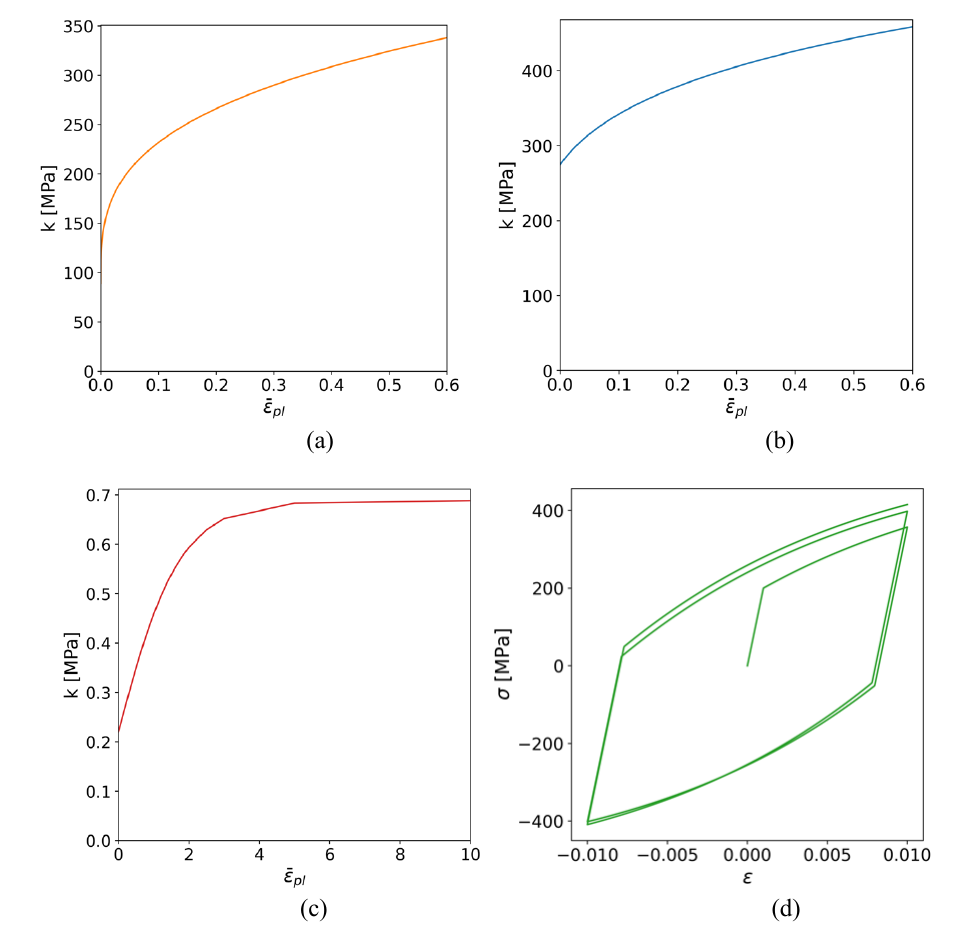}
  \caption{Hardening laws: isotropic hardening of (a) copper, (b) dual phase steel, and (c) metallic foam. (d) Combined kinematic-isotropic hardening response of low carbon steel for cyclic tension-compression loading}
  \label{fig:2}
\end{figure}

Following \citet{bonatti2021}, the architectures are trained, validated, and tested
on data representative of the history-dependent behavior of four distinct materials:

\begin{enumerate}
\item[(1)] Copper: $J_2$-plasticity with isotropic strain hardening, in which the
deformation resistance $k$ evolves with the equivalent plastic strain
$\bar{\varepsilon}_p$ as
\begin{equation}\label{eq:25}
k\!\left[\bar{\varepsilon}_p\right] = B + C\!\left(\bar{\varepsilon}_p\right)^{n}
\end{equation}
with the parameters $B = 89\,\mathrm{MPa}$, $C = 292\,\mathrm{MPa}$, and $n = 0.31$
(Fig.~\ref{fig:2}a). The isotropic elastic behavior is specified by the Young's
modulus $E = 124\,\mathrm{GPa}$ and the Poisson's ratio $\nu = 0.34$; the model
features six external state variables (stress-tensor components) and one internal
state variable (equivalent plastic strain).

\item[(2)] Dual Phase (DP) steel: $J_2$-plasticity with isotropic Swift hardening
where the deformation resistance $k$ evolves as a function of the equivalent plastic
strain $\bar{\varepsilon}_p$ according to
\begin{equation}\label{eq:26}
k\!\left[\bar{\varepsilon}_p\right]
  = A\!\left(\varepsilon_0 + \bar{\varepsilon}_p\right)^{n}
\end{equation}
with the parameters $A = 500\,\mathrm{MPa}$, $\varepsilon_0 = 0.05$, and $n = 0.2$
(Fig.~\ref{fig:2}b). The isotropic elastic behavior is specified by the Young's
modulus $E = 200\,\mathrm{GPa}$ and the Poisson's ratio $\nu = 0.3$; the model
features the same number of state variables as the copper materials.

\item[(3)] Metallic foam: Deshpande-Fleck type model with isotropic hardening
(Fig~\ref{fig:2}c), a uniaxial-to-hydrostatic yield stress ratio of $k = 1.1$, and a
plastic Poisson's ratio of $\nu_p = 0$ (Fig.~\ref{fig:2}d). The isotropic elastic
behavior is specified by the Young's modulus $E = 3\,\mathrm{MPa}$ and the elastic
Poisson's ratio $\nu = 0.2$; though its yield surface definition is different from
the copper and DP steel models, it features the same number of state variables.

\item[(4)] Low Carbon (LC) steel: $J_2$-plasticity with combined
isotropic-kinematic hardening, composed of Voce-type isotropic hardening
\begin{equation}\label{eq:27}
k\!\left[\bar{\varepsilon}_p\right]
  = k_0 + Q\!\left(1 - \exp\!\left(-\beta\bar{\varepsilon}_p\right)\right)
\end{equation}
with the parameters $k_0 = 200\,\mathrm{MPa}$, $Q = 2\,\mathrm{GPa}$ and
$\beta = 0.26$, and kinematic hardening defining the evolution of the back
stress-tensor $\bm{\alpha}$ as
\begin{equation}\label{eq:28}
d\alpha = \frac{C}{k_0}(\bm{\sigma} - \bm{\alpha})\,\mathrm{d}\bar{\varepsilon}_p
  - \gamma\bm{\alpha}\!\left(d\bar{\varepsilon}_p\right)
\end{equation}
with the parameters $C = 25.5\,\mathrm{GPa}$ and $\gamma = 81$. The isotropic
elastic behavior is specified by the Young's modulus $E = 210\,\mathrm{GPa}$ and the
Poisson's ratio $\nu = 0.3$; Fig.~\ref{fig:2}d elucidates the material's cyclic
stress-strain response for uniaxial stress loading. In addition to the seven state
variables of an isotropic hardening model, the kinematic hardening adds five
additional internal state variables (back stress-tensor components), leading to a
total of 12 state variables.
\end{enumerate}

\subsection{Strain path sampling}\label{sec:3.2}

The six-dimensional strain space is explored via random walks within prescribed
bounds. Each path comprises 126 data points and is generated as follows:

\begin{enumerate}
\item A main path is constructed by randomly sampling five strain tensors
$\bm{\varepsilon}^{(i)}$ whose deviatoric and spherical parts satisfy
\begin{equation}\label{eq:29}
\sqrt{\frac{2}{3}\,dev(\bm{\varepsilon}^{(i)})\!:\!dev(\bm{\varepsilon}^{(i)})}
  \leq \varepsilon_{max}
\end{equation}
and
\begin{equation}\label{eq:30}
\left[tr(\bm{\varepsilon}^{(i)})\right] \leq \varepsilon_{tr}^{max}\,,
\end{equation}
with $\varepsilon_{dev}^{max} = 0.07$ and $\varepsilon_{tr}^{max} = 0.1$. Starting
from the origin $\bm{\varepsilon}^{(0)} = \bm{0}$, the main path consists of the
sequence $\{\bm{\varepsilon}^{(0)},\bm{\varepsilon}^{(25)},
\bm{\varepsilon}^{(50)},\bm{\varepsilon}^{(75)},\bm{\varepsilon}^{(100)},
\bm{\varepsilon}^{(125)}\}$.

\item Subsequently, each straight segment of the main path is uniformly subdivided
into five sub-segments, yielding a refined path of 26 points.

\item Noise $\delta\varepsilon = \eta_1\eta_2$ is added to the newly-created
intermediate points, where the magnitude $\eta_1$ is sampled log-uniformly from the
interval $[0.002,0.05]$, and modulated by a rugosity factor $\eta_2$ sampled
uniformly from the interval $[0,1]$; the noise direction $\bm{n}$ is uniformly drawn
from a unit hyperball in $\mathbb{R}^6$. The strain bounds are enforced throughout
by repeated sampling if necessary.

\item Each of the 25 segments of the rough path is subdivided further into five
subsegments, yielding the final path $\{\bm{\varepsilon}^{(0)},
\bm{\varepsilon}^{(1)},...,\bm{\varepsilon}^{(125)}\}$ with 125 segments.
\end{enumerate}

\subsection{Dataset construction}\label{sec:3.3}

Single-element simulations are performed with the finite element software
Abaqus/explicit to compute the stress sequence $\{\bm{\sigma}^{(0)},
\bm{\sigma}^{(1)},...,\bm{\sigma}^{(125)}\}$ for a given strain path sequence
$\{\bm{\varepsilon}^{(0)},\bm{\varepsilon}^{(1)},...,\bm{\varepsilon}^{(125)}\}$.
The strain paths are converted into nodal displacements and applied to a single
hexahedral element (C3D8R). From each simulation, both the Hencky strain-tensor
history $\bm{\varepsilon}[t]$ and the Cauchy stress-tensor history $\bm{\sigma}[t]$
are extracted at the central integration point.

\begin{figure}[H]
  \centering
  \includegraphics[width=\linewidth]{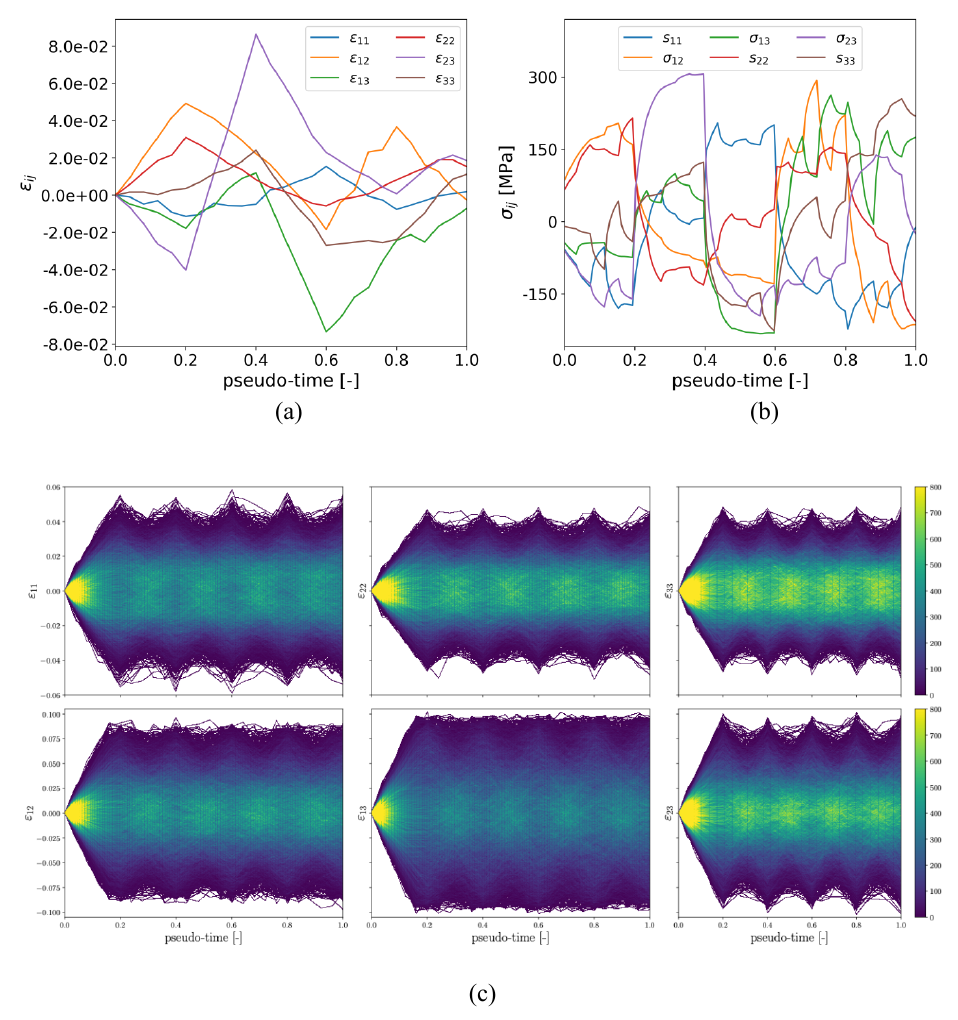}
  \caption{Training data: (a) example of a random walk in strain space, (b) corresponding sequence in stress space for the mixed hardening materials, (c) superposition of all 10'000 strain paths comprised in the dataset with colormap highlighting the frequency of individual points.}
  \label{fig:3}
\end{figure}

For each material, a dataset is generated composed of 10,000 sequence pairs,
describing the mapping of a strain-tensor increment sequence
$\{\bm{\Delta\varepsilon}^{(1)},\bm{\Delta\varepsilon}^{(2)},...,
\bm{\Delta\varepsilon}^{(125)}\}$, with $\bm{\Delta\varepsilon}^{(i)} =
\bm{\varepsilon}^{(i)} - \bm{\varepsilon}^{(i-1)}$, to a stress-tensor sequence
$\{\bm{\sigma}^{(1)},...,\bm{\sigma}^{(125)}\}$. Figure~\ref{fig:3}a depicts the
history of the six strain-tensor components of an example loading path. The
corresponding stress-tensor component histories are shown in Fig.~\ref{fig:3}b.
They feature sharp peaks which are attributed to reloading-unloading along different
directions. Figure~\ref{fig:3}c visualizes the distribution of all 6-D loading paths
comprised in the dataset with 2-D heatmaps. Although no individual strain component
is directly bounded, the bounds imposed via Eqs.~\eqref{eq:29} and \eqref{eq:30}
keep the normal and shear components within approximately $[-0.06, 0.06]$ and
$[-0.09, 0.09]$, respectively. Due to the continuity of the strain paths, values
near zero occur most frequently, reflecting repeated load reversals.

Of the 10,000 sequence-sequence pairs, 8,000 are used for training, 1,000 for
validation, and 1,000 for testing.

%% file: sections/04-results.tex
\section{Results}\label{sec:4}

The computational assessment is designed to distinguish representational accuracy from
the numerical and algorithmic properties of the constitutive update. To this end, the models
are evaluated at the material-point level under prescribed strain histories. This setting
isolates the intrinsic behavior of the learned constitutive update from mesh dependence,
global equilibrium iterations, contact treatment, and other solver-specific effects.
Importantly, it enables controlled variations of the strain-path discretization while keeping
both the underlying loading path and the reference constitutive response unchanged, thereby
providing a direct assessment of resolution sensitivity. Independent hyperparameter studies
are first conducted for the MSC and CSS architectures (Section~\ref{sec:4.1}), followed by a
comparison of stress-prediction accuracy (Section~\ref{sec:4.2}).
Section~\ref{sec:4.3} then examines numerical consistency through stationarity and
sensitivity to strain-path resolution. The computational benefit of the parallelizable
state-space recurrence is quantified through training-time scaling with sequence length
(Section~\ref{sec:4.4}), while Section~\ref{sec:4.5} evaluates the amount of strain-stress
data required to reach a given accuracy. Finally, Section~\ref{sec:4.6} probes whether the
learned CSS state retains physically meaningful structure.

All results are evaluated on the held-out testing dataset unless stated otherwise. For a
given stress path with $n_t$ time steps, the prediction quality for a stress component
$\sigma_{ij}$ is quantified by the Mean Squared Error (MSE). The Normalized Root Mean
Squared Error (NRMSE), adapted from \citet{ferreira2025}, is also reported,

\begin{equation}\label{eq:31}
\mathrm{NRMSE}\left(\sigma_{ij}\right)
= \frac{\mathrm{RMSE}\left(\sigma_{ij}\right)}
       {\mathrm{MAV}\left(\sigma_{ij}\right)}, \quad
\{i,j\} = 1,\ldots,n_{\mathrm{dim}},
\end{equation}

where $\mathrm{RMSE}\left(\sigma_{ij}\right)$ and $\mathrm{MAV}\left(\sigma_{ij}\right)$
denote the Root Mean Squared Error and the Mean Absolute Value of the stress component
$\sigma_{ij}$ over all $n_t$ time steps, respectively. The NRMSE is an easily interpretable,
dimensionless relative error. When reporting the performance over the full testing dataset,
the NRMSE of each stress component is averaged across all stress paths.

\subsection{Hyperparameter study}\label{sec:4.1}

\subsubsection{Hyperparameter study for the MSC model}\label{sec:4.1.1}

A grid search is conducted to identify the optimal MSC model architecture for each material.
All three hyperparameters are varied: (i) number of state variables,
$s \in [1, 2,\ldots, 40]$, (ii) depth of the internal layers, $d \in [2, 4, 6, 8]$, and
(iii) width of the internal layers, $w \in [15, 25, 40, 50, 80]$. The resulting models span
between 894 and 100,040 trainable parameters. The grid search comprises 312 trials, each
trained with the ADAM optimizer for 1,500 epochs using a batch size of 64, an initial
learning rate $\mu_0 = 5 \times 10^{-3}$, a learning-rate decay pre-factor $\eta = 0.03$, a
reload patience of 20 epochs, and a maximum gradient norm of 0.1.

\begin{figure}[H]
  \centering
  \includegraphics[width=\linewidth]{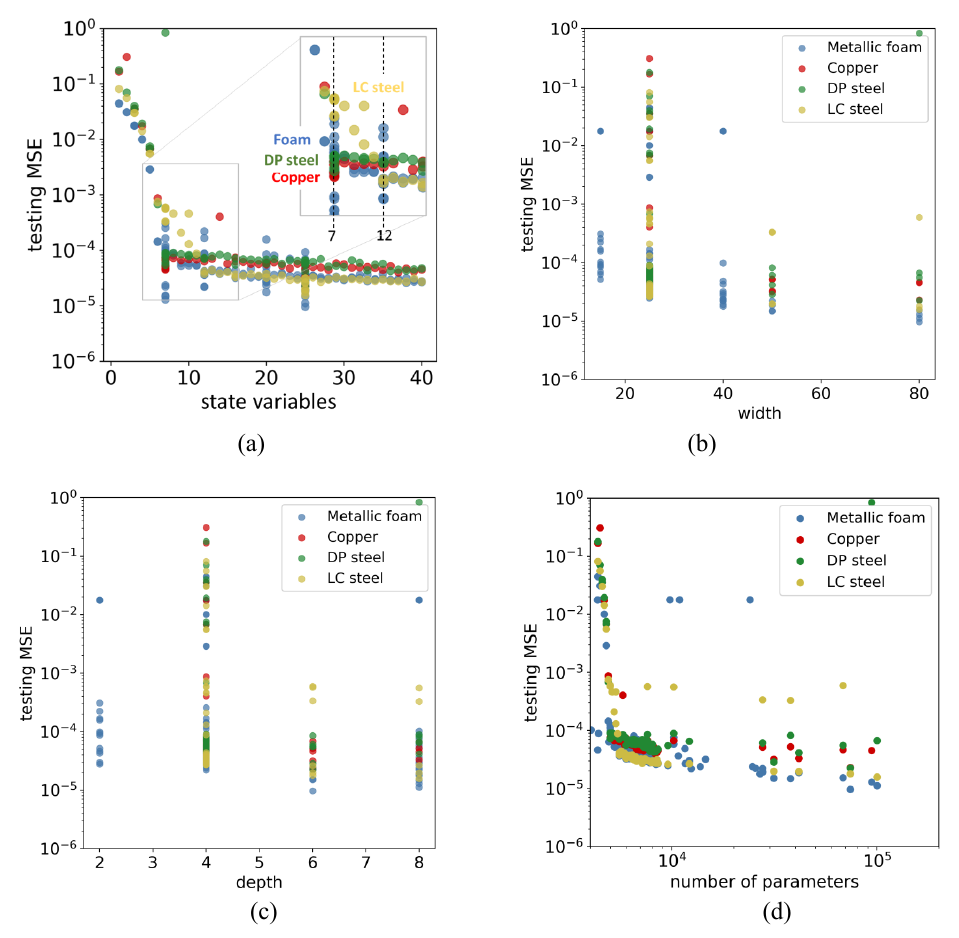}
  \caption{MSC hyperparameter study. Effects of (a) number of state variables, (b) width of internal layers, (c) depth of internal layers, and (d) total number of parameters.}
  \label{fig:4}
\end{figure}

Figure~\ref{fig:4} summarizes the results from the grid search. The metallic foam attains the
lowest validation MSE ($9.64 \times 10^{-6}$), followed by the LC steel
($1.57 \times 10^{-5}$), the DP steel ($2.25 \times 10^{-5}$), and the copper
($2.29 \times 10^{-5}$). Among the hyperparameters, the number of state variables $s$
(Fig.~\ref{fig:4}a) has the strongest influence on the model's generalization ability.
Performance degrades sharply for small $s$, while increasing $s$ beyond 7 offers only limited
benefits for foam, DP steel and copper. The exception is the LC steel (mixed hardening
material), where the MSE continues to decrease until $s = 12$. This behavior is consistent
with the minimal-state character of the MSC model, since these transition values match to the
number of state variables in the respective physics-based models.

The number of trainable parameters scales with the width of the MSC's internal encoder. The
corresponding MSE vs.\ width plot (Fig.~\ref{fig:4}b) shows performance improving from a
median MSE of $1.58 \times 10^{-4}$ at $w = 15$ to $2.27 \times 10^{-5}$ at $w = 80$. The
number of parameters also scales with the encoder depth, but comparing the medians for
different depths reveals a non-monotonic relationship (Fig.~\ref{fig:4}c): median MSE values
of $9.8 \times 10^{-5}$, $5.2 \times 10^{-5}$ and $4.8 \times 10^{-5}$ are obtained for
$d = 2$, 4 and 8, respectively, while $d = 6$ reaches the best median of
$3.2 \times 10^{-5}$.

Larger models generally perform better, though with substantial variance (see for example the
results for the copper material in Fig.~\ref{fig:4}d). Models with 5,000--10,000 parameters
achieve a median MSE of $4.5 \times 10^{-5}$, while those with 50,000--100,000 parameters
reach $2.3 \times 10^{-5}$. The best configurations cluster around 70,000--100,000
parameters, though compact models with approximately 25 state variables and 30,000 parameters
(e.g., $s = 25$, $d = 6$, $w = 50$) offer a roughly threefold reduction in model size at a
modest accuracy cost (MSE $= 1.50 \times 10^{-5}$). For the remainder of the paper, we retain
the following MSC configurations as ``best'' models:

\begin{itemize}
  \item Copper and DP steel: $p = 30$, $h = 60$, $d = 6$, about 67k parameters,
  \item Low carbon steel: $p = 40$, $h = 60$, $d = 8$, about 105k parameters,
  \item Metallic foam: $p = 30$, $h = 60$, $d = 8$, about 89k parameters.
\end{itemize}

\subsubsection{Hyperparameter study for the CSS model}\label{sec:4.1.2}

The hyperparameters of the CSS model are (i) the latent input dimension $h$, (ii) the number
$d$ of S5-blocks, and (iii) the complex state dimension $p$ of each block. The results are
reported in terms of the number of real-valued state variables $s = 2pd$, with the factor 2
accounting for the complex-valued entries of the state vectors $\bm{\chi}$. All
configurations are trained for up to 3,000 epochs on an RTX 4090 GPU.

\begin{figure}[H]
  \centering
  \includegraphics[width=\linewidth]{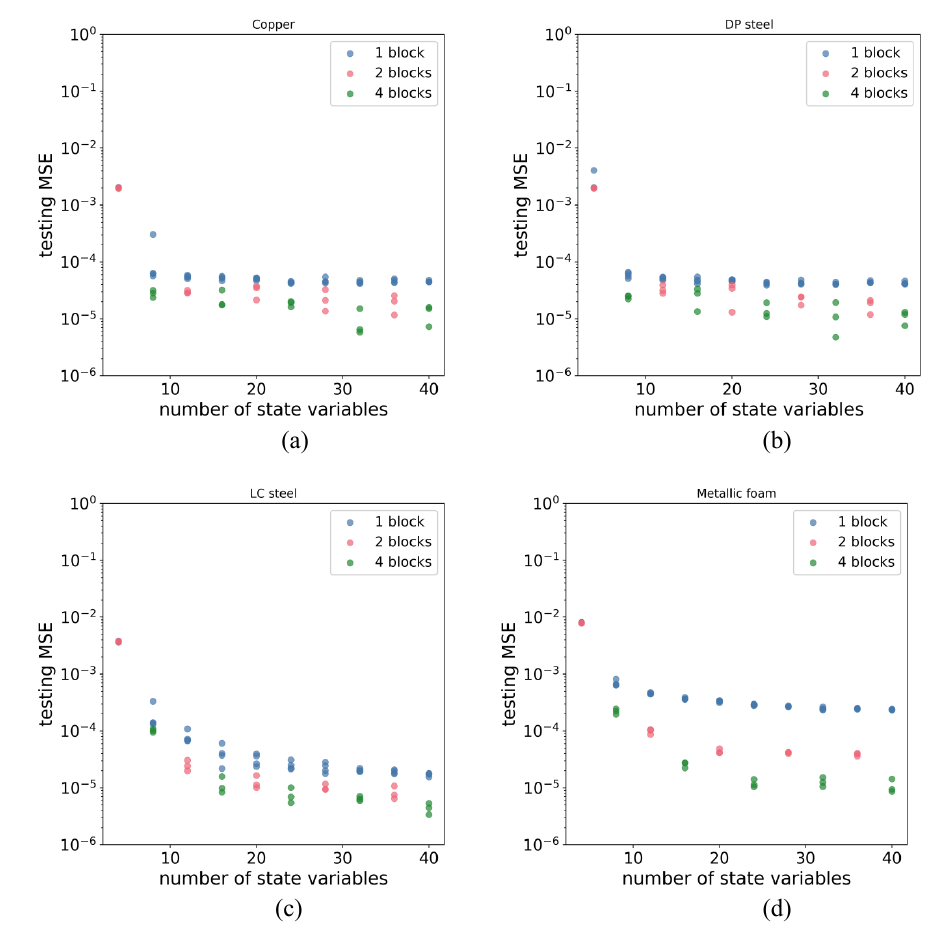}
  \caption{CSS hyperparameter study \#1 -- effect of the number of real-valued state variables: (a) copper, (b) DP steel, (c) low-carbon steel, (d) metallic foam.}
  \label{fig:5}
\end{figure}

In a first parameter study, $p \in [1,\ldots,20]$ and $d \in [1, 2, 4]$ are considered,
resulting in a number of state variables $s$ ranging from 2 to 160. Furthermore, the latent
input dimension $h$ is varied such that the total parameter count remains within
15,000-34,000. Figure~\ref{fig:5} shows the validation MSE as a function of the state
dimension $s$. For $d = 1$ (blue dots), we define models with $s = \{4, 8, 12, 16, ..\}$,
while for $d = 2$ (red dots) and $d = 4$ (green dots), we have
$s = \{4, 12, 20, \ldots\}$ and $s = \{8, 16, 24, \ldots\}$, respectively.

When looking at the results for the models with $d = 1$ for the copper
(Fig.~\ref{fig:5}a), DP steel (Fig.~\ref{fig:5}b), and metallic foam (Fig.~\ref{fig:5}d) --
whose data-generating physics-based models all feature seven state variables) -- the
validation MSE improves substantially when increasing the number of state variables up to
$s = 8$. Beyond that point, further increasing the length of the state vector is less
beneficial. For the LC steel (Fig.~\ref{fig:5}c) -- whose data-generating model featured 12
state variables) -- the improvements in the CSS's performance tend to stagnate when
increasing the number of state variables beyond 16 (Fig.~\ref{fig:5}c). In other words,
similar to the MSC, the CSS model appears able to implicitly infer the number of required
state variables from seeing the training data.

\begin{figure}[H]
  \centering
  \includegraphics[width=\linewidth]{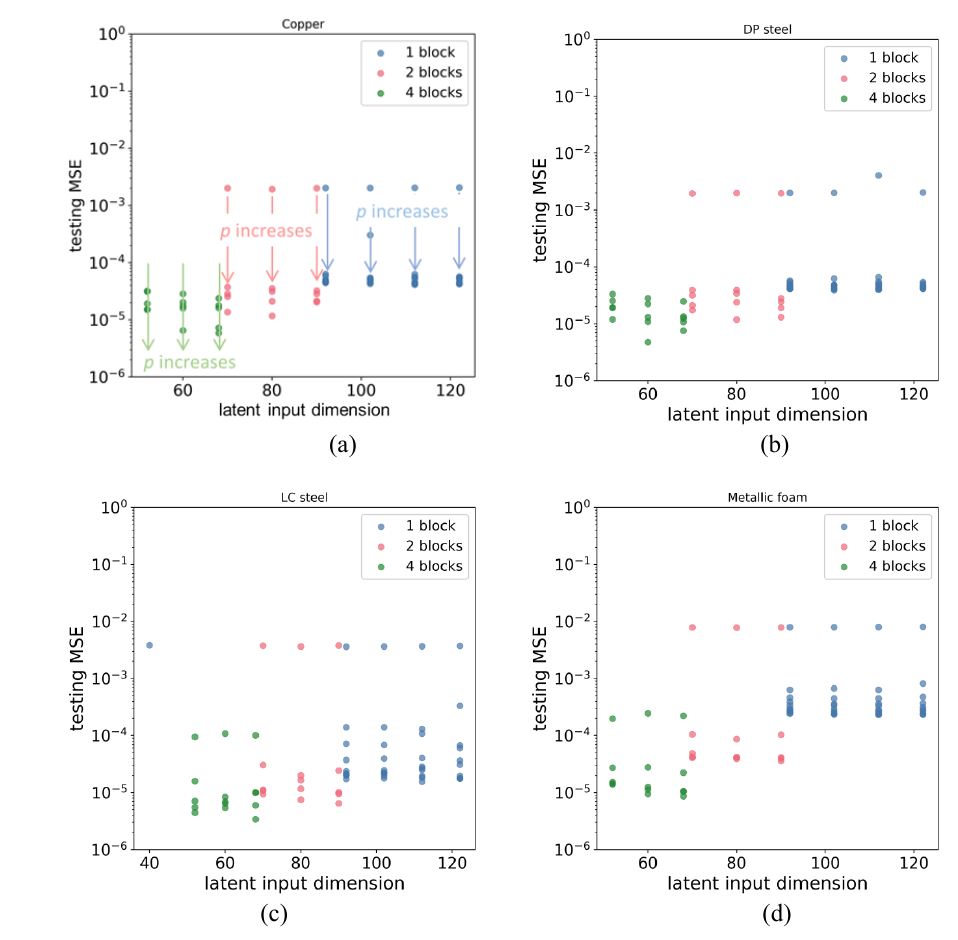}
  \caption{CSS hyperparameter study \#1 -- effect of the latent input dimension: (a) copper, (b) DP steel, (c) low-carbon steel, (d) metallic foam.}
  \label{fig:6}
\end{figure}

Increasing the number of state blocks while holding the number of state variables fixed seems
to be beneficial for all four materials. In particular, for the foam material, the prediction
accuracy increases by more than one order of magnitude when using four blocks instead of one
(Fig.~\ref{fig:5}d). When plotting the same data against the latent input dimension $h$
(Fig.~\ref{fig:6}), the MSE improves only mildly with $h$, at least in the examples
considered under the constraint of keeping the overall parameter count within the bounds of
15,000 to 34,000. The highest-MSE outliers correspond to models with an insufficient number
of state variables. The effect of the number of blocks appears to be dominant: models with
$d = 4$ consistently achieve lower MSE than $d = 2$ or $d = 1$, even with a smaller latent
input dimension. If the latter is frozen, a wide spread remains due to the underlying
variation in the complex state dimension $p$; varying $p$ has a monotonic effect: the larger
its value, the lower the attained MSE.

\begin{figure}[H]
  \centering
  \includegraphics[width=\linewidth]{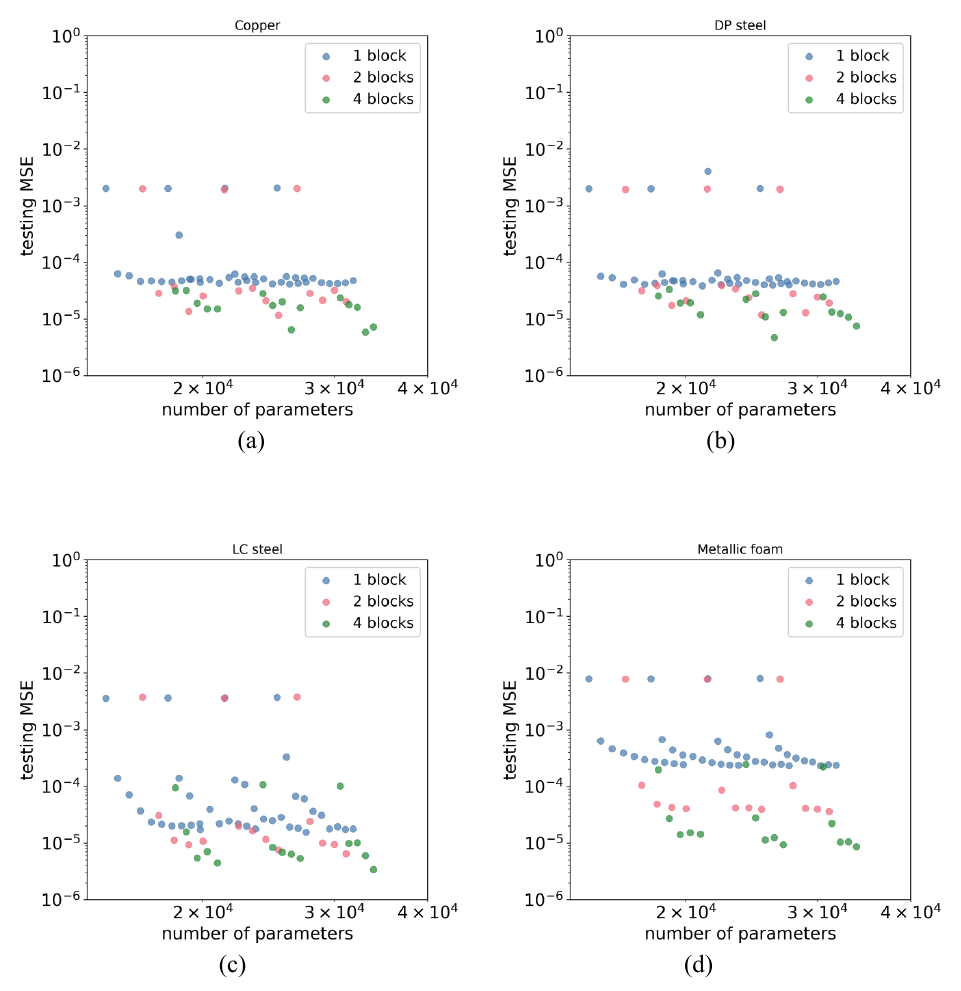}
  \caption{CSS hyperparameter study \#1 -- effect of the total number of parameters: (a) copper, (b) DP steel, (c) low-carbon steel, (d) metallic foam.}
  \label{fig:7}
\end{figure}

These conclusions may also be drawn from the plots of the MSE against the number of trainable
parameters (Fig.~\ref{fig:7}). For the metallic foam (Fig.~\ref{fig:7}d), four monotonically
decreasing curves are observed for $d = 1$, each corresponding to a different latent input
dimension $h$. For $h = 90$ (leftmost curve), the MSE decreases as $p$ increases, which
results in an increase of the total parameter count. A similar behavior is observed for the
curve branches for $h = 100, 110$, and $120$ (rightmost curve), while major improvements are
achieved when exceeding the minimum number of state variables required and by increasing the
number of S5 blocks. Overall, the model accuracy improves with an increase in any of the
three architecture-defining hyperparameters: above a critical minimum of $p$, $d$ exerts the
strongest influence.

\begin{figure}[H]
  \centering
  \includegraphics[width=\linewidth]{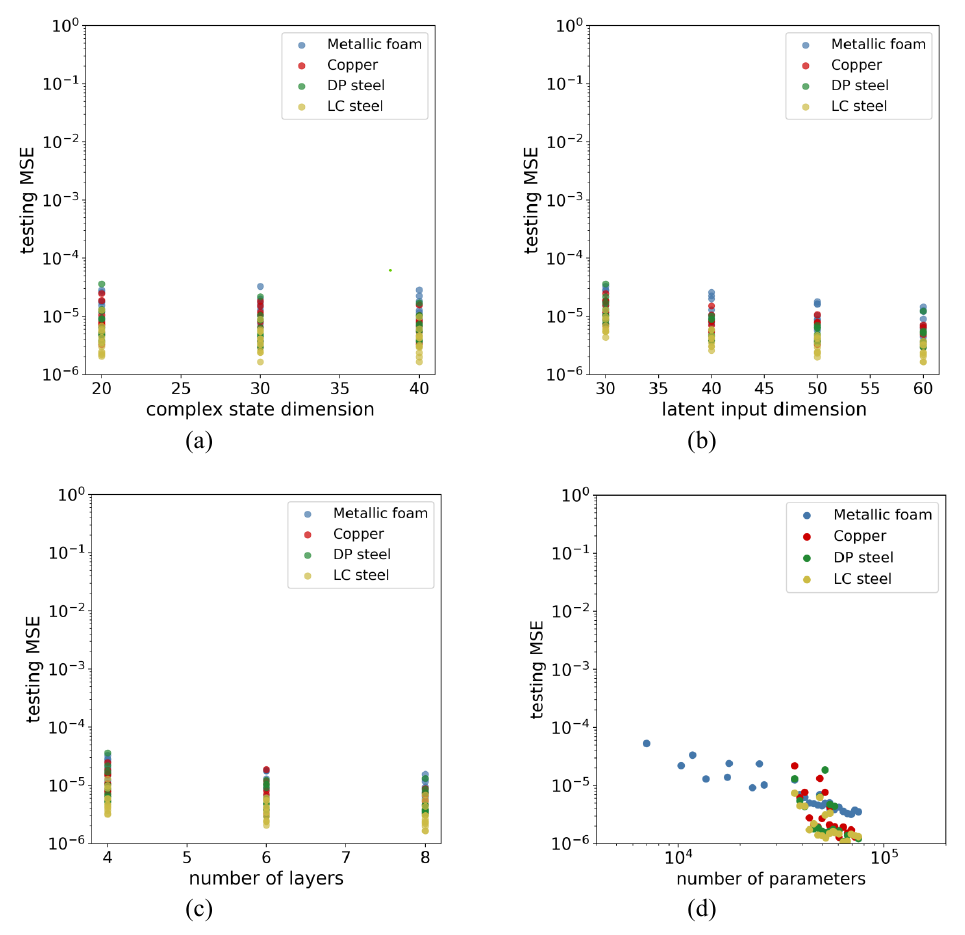}
  \caption{CSS hyperparameter study \#2. Effects of (a) complex state dimension, (b) latent input dimension, (c) number of S5 blocks, (d) total number of trainable parameters.}
  \label{fig:8}
\end{figure}

A second search explores significantly larger models to probe the model's learning ability.
It sweeps long latent input dimensions $h \in [30, 40, 50, 60]$, large complex state
dimensions, $p \in [20, 30, 40]$, and high numbers of stacked S5 blocks, $d \in [4, 6, 8]$,
leading to models with up to 105k parameters. Figure~\ref{fig:8} shows the resulting MSE
trends for the different materials. As observed previously, increasing the complex state
dimension beyond the material-dependent minimum requirement is little beneficial
(Fig.~\ref{fig:8}a). A state dimension of $p = 30$ provides the computational optimum for all
materials, except for the low carbon steel (mixed hardening material), where $p = 40$ yields
the lowest MSE in the above parameter space. Irrespective of the material, the largest latent
input dimension ($h = 60$) consistently yields the best results, albeit with limited
improvements over models with $h = 30$ (Fig.~\ref{fig:8}b). More heterogeneous results are
found when varying the number of blocks (Fig.~\ref{fig:8}c), advocating six blocks for the
copper and DP steel, versus eight blocks for the metallic foam and mixed-hardening LC steel.
Overall, a strong correlation is observed between the test MSE and the total number of
parameters (Fig.~\ref{fig:8}d). In logarithmic space, the test loss decreases approximately
linearly with increasing parameter count, indicating excellent scaling behavior of the CSS
model.

The ``best'' models that have been included in our hyperparameter study are:

\begin{itemize}
  \item Copper and DP steel: $p = 30$, $h = 60$, $d = 6$, about 67k parameters,
  \item Low carbon steel: $p = 40$, $h = 60$, $d = 8$, about 105k parameters,
  \item Metallic foam: $p = 30$, $h = 60$, $d = 8$, about 89k parameters.
\end{itemize}

\subsection{Comparison of the MSC and CSS model prediction accuracy}\label{sec:4.2}

\begin{figure}[H]
  \centering
  \includegraphics[width=\linewidth]{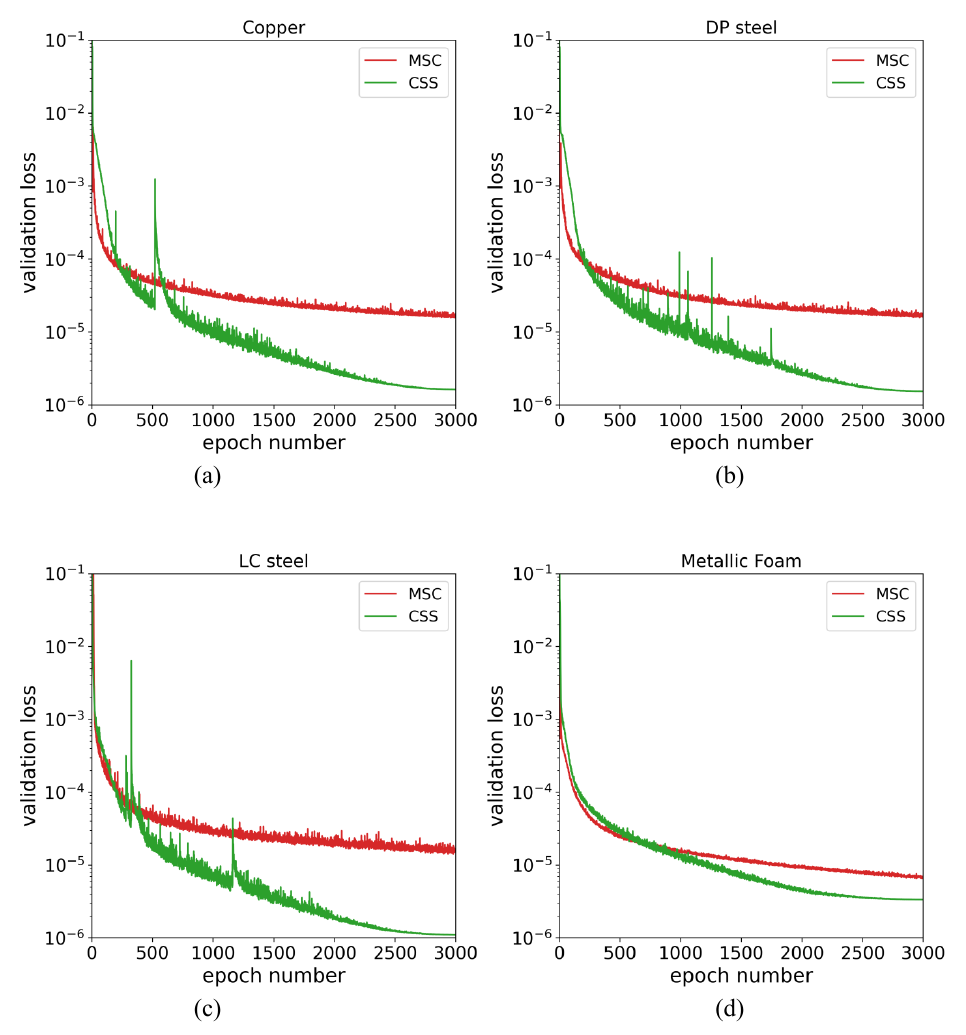}
  \caption{Performance of best MSC and CSS models on different datasets. Validation loss over training epochs for all materials: (a) copper, (b) DP steel, (c) low-carbon steel, (d) metallic foam.}
  \label{fig:9}
\end{figure}

The best MSC and CSS models are retrained using the same batch size of 64 and the same number
of epochs (up to 3,000). The resulting performance curves, i.e., the validation loss vs.\
epoch number, are shown for all four materials in Fig.~\ref{fig:9}. For the three plastically
incompressible materials (Figs.~\ref{fig:9}a-c), the CSS systematically outperforms the MSC
by approximately one order of magnitude, attaining validation losses around $10^{-6}$. For
the metallic foam (Fig.~\ref{fig:9}d), both models converge to a loss near $10^{-5}$.

\begin{figure}[H]
  \centering
  \includegraphics[width=\linewidth]{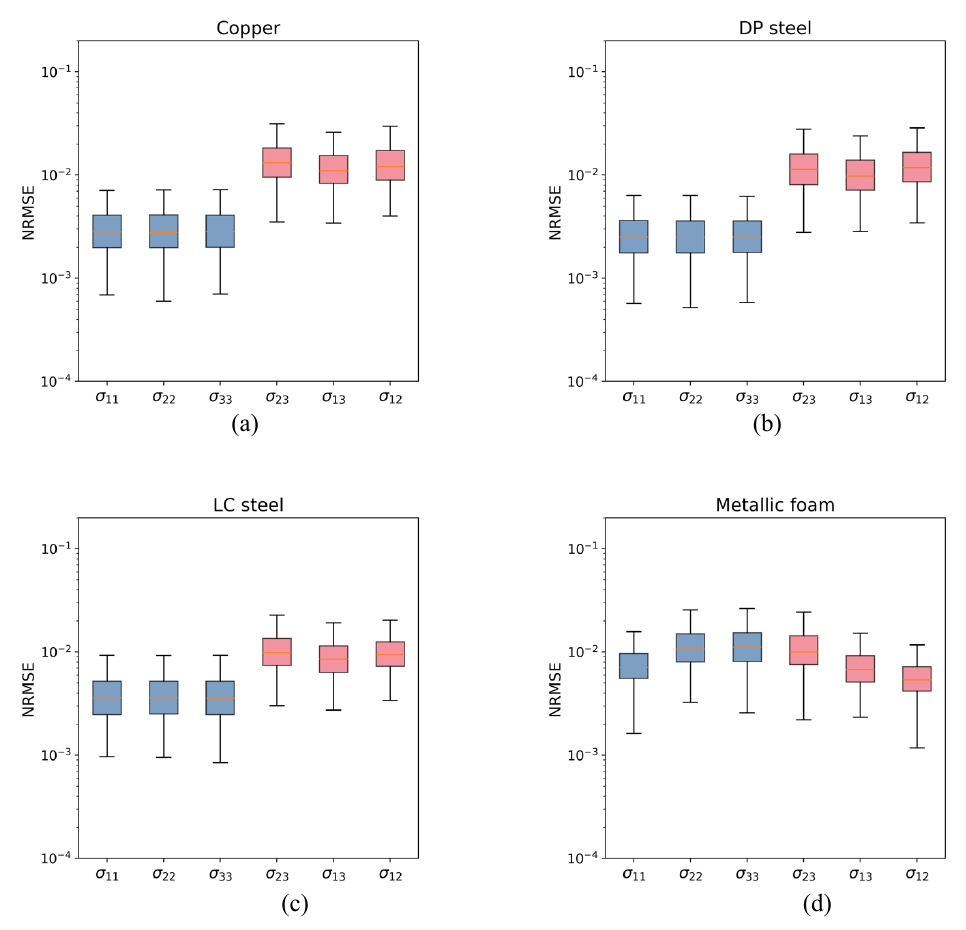}
  \caption{Performance of best MSC models on different datasets. Normalized Root Mean Square Error (NRMSE) for all stress tensor components for all materials: (a) copper, (b) DP steel, (c) low carbon steel, (d) metallic foam.}
  \label{fig:10}
\end{figure}

\begin{figure}[H]
  \centering
  \includegraphics[width=\linewidth]{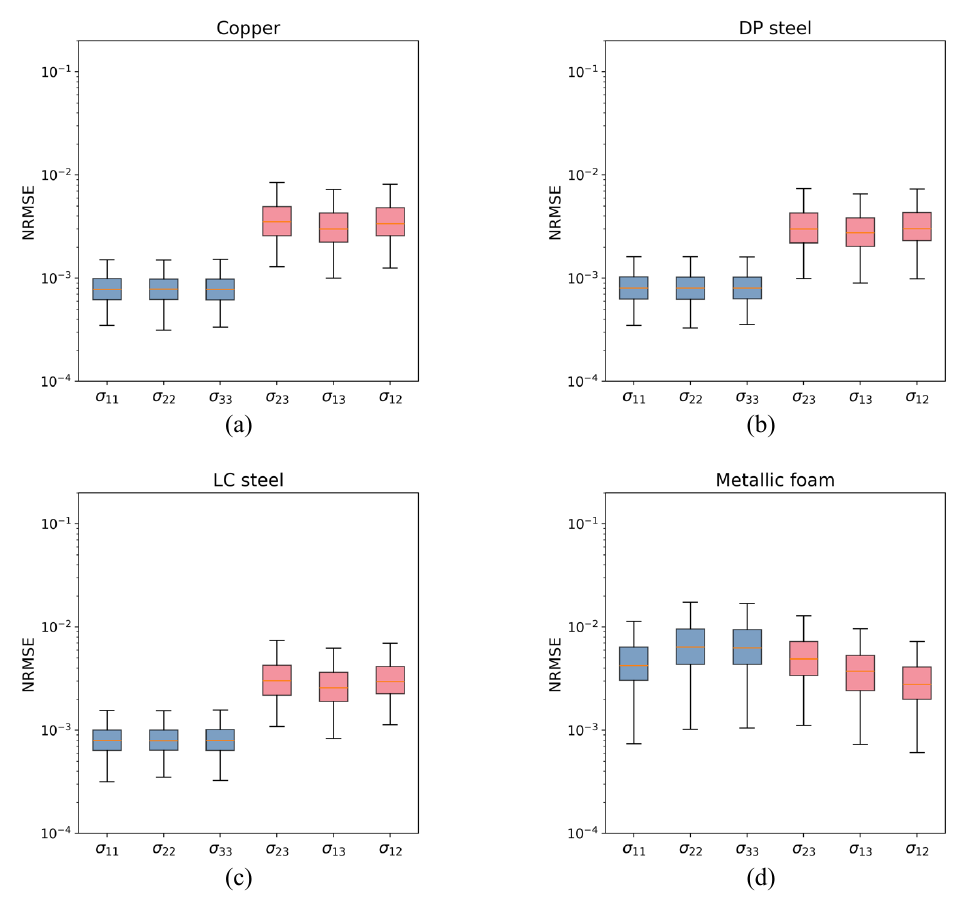}
  \caption{Performance of best CSS models on different datasets. Normalized Root Mean Square Error (NRMSE) for all stress tensor components for all materials: (a) copper, (b) DP steel, (c) low carbon steel, (d) metallic foam.}
  \label{fig:11}
\end{figure}

The component-wise NRMSE distributions for the MSC models are shown in Fig.~\ref{fig:10}. For
the metallic foam (Fig.~\ref{fig:10}d), relative errors are approximately 1\% across all
stress-tensor components. For the other materials (Figs.~\ref{fig:10}a-c), the normal stress
components systematically achieve lower relative errors than the shear components, likely due
to their larger absolute magnitudes associated with plastic incompressibility. Notably, the
NRMSE profiles are qualitatively similar across all three von Mises materials despite
differences in yield strength and hardening behavior. Figure~\ref{fig:11} shows the
corresponding NRMSE distributions for the trained CSS models. Irrespective of the material at
hand, the CSS always achieves low errors, with medians between $10^{-3}$ and $10^{-2}$.

\begin{figure}[H]
  \centering
  \includegraphics[width=\linewidth]{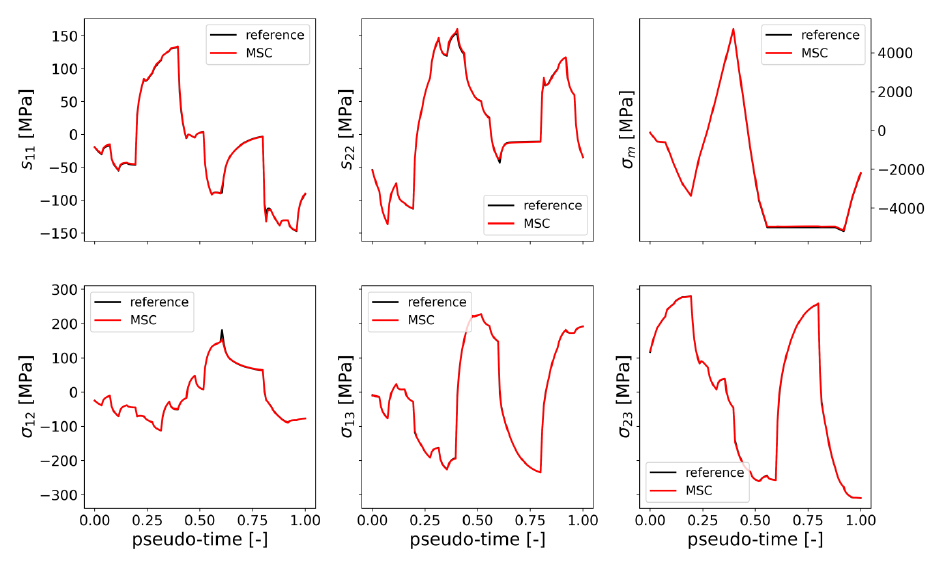}
  \caption{Testing of the MSC model for low carbon steel (mixed isotropic-kinematic hardening): Stress predictions for the 95\%-quantile path (high error) in the testing dataset. The black reference curve is obtained through a single-element FE analysis using the data-generating constitutive model. Instead of reporting all three normal stress components, we plot two deviatoric components and the hydrostatic stress.}
  \label{fig:12}
\end{figure}

\begin{figure}[H]
  \centering
  \includegraphics[width=\linewidth]{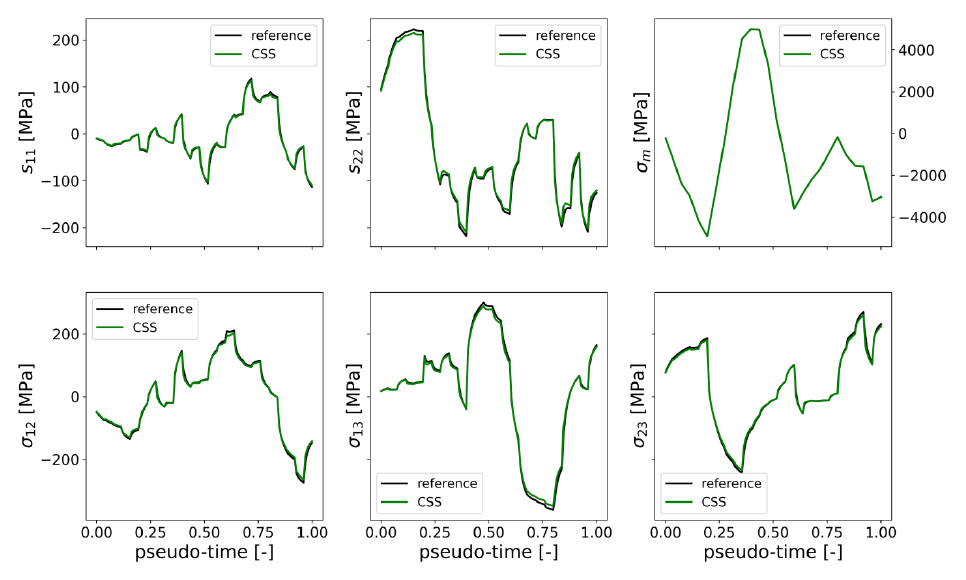}
  \caption{Testing of the CSS model for low carbon steel (mixed isotropic-kinematics hardening): Stress predictions for the 95\%-quantile path (high error) in the testing dataset. The black reference curve is obtained through a single-element FE analysis using the data-generating constitutive model. Instead of reporting all three normal stress components, we plot two deviatoric components and the hydrostatic stress.}
  \label{fig:13}
\end{figure}

To provide an engineering interpretation of the error levels achieved, the model predictions
for the low carbon steel are compared against the ground truth for a worst-case loading
95\%-percentile test path (Figs.~\ref{fig:12} and~\ref{fig:13}). Despite its relatively high
MSE, the MSC predictions (red curves) show satisfactory agreement with the ground truth
(black curves) from an engineering perspective (Fig.~\ref{fig:12}). Slight deviations of a
few MPa become visible towards the end of the sequence, which is characteristic of recurrent
networks trained on long sequences. Similarly, Fig.~\ref{fig:13}, comparing the black curves
(ground truth) and the green curves (CSS predictions), confirms the good accuracy of the CSS
model.

\subsection{Numerical consistency and resolution robustness}\label{sec:4.3}

\subsubsection{Stationarity}\label{sec:4.3.1}

\begin{figure}[H]
  \centering
  \includegraphics[width=0.8\linewidth]{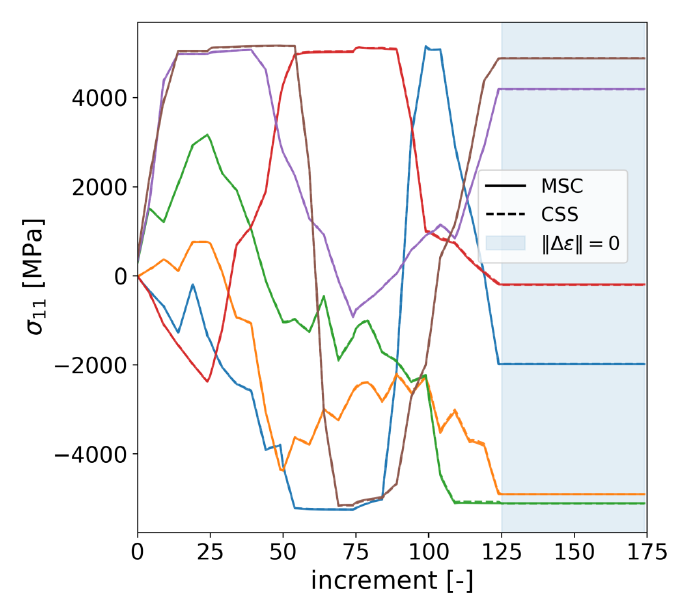}
  \caption{Stationarity check for the MSC and CSS models: 50 zero-increment steps are appended to random strain paths from the testing dataset, and no stress drift is reported (only 11-component shown for better visualization).}
  \label{fig:14}
\end{figure}

By construction, both the MSC and CSS models satisfy the stationarity condition exactly: when
a zero-norm strain increment is applied, the stress remains unchanged. This is verified
empirically by appending 50 zero-increment steps to random strain paths from the testing
dataset and recording any stress drift. Figure~\ref{fig:14} shows the $\sigma_{11}$ component
over the appended interval for several representative paths. As expected, the predictions by
both architectures maintain constant stress throughout the zero-increment interval, thereby
perfectly satisfying the stationarity requirement.

\subsubsection{Resolution invariance}\label{sec:4.3.2}

Resolution invariance is assessed by training both models after changing the discretization of
the training sequences through resampling. The original FE-generated dataset comprises 125
linear strain-path segments, with the stress retained at the corresponding 125 segment
endpoints. To evaluate the numerical behavior at finer resolutions, additional datasets are
generated by linear interpolation, without rerunning finite element simulations. Specifically,
datasets with 250, 500, 1,000, 2,000, and 10,000 increments are generated through repeated
segment division. Models are trained and evaluated on combinations of training and testing
resolutions.

\begin{figure}[H]
  \centering
  \includegraphics[width=0.62\linewidth]{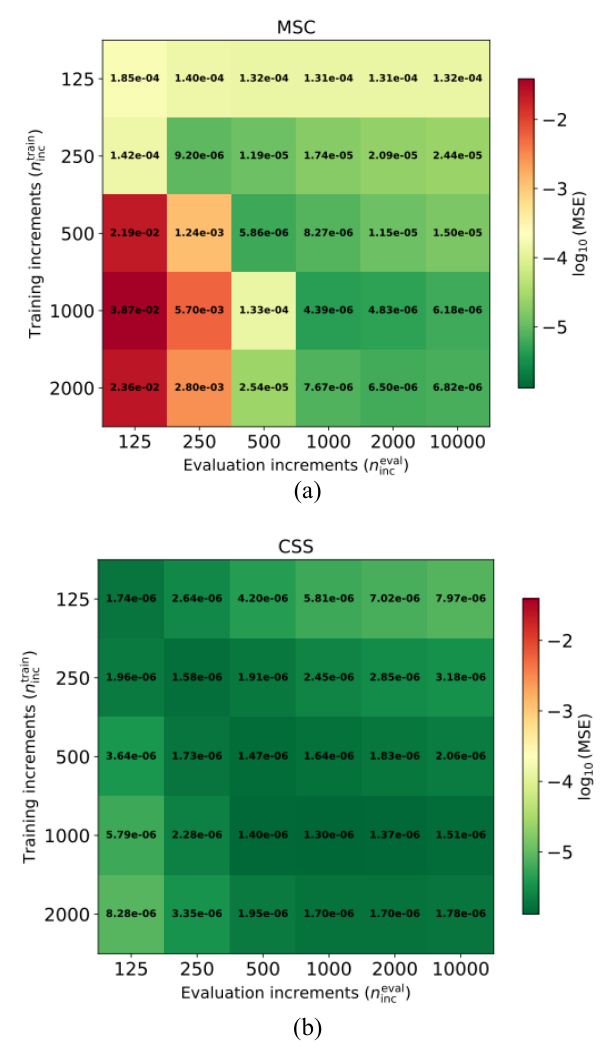}
  \caption{Testing MSE for models trained and tested at different strain path resolutions. A high number of increments (per path) corresponds to a small strain increment size: (a) MSC models, (b) CSS models.}
  \label{fig:15}
\end{figure}

The resulting test MSE heatmap for the MSC (Fig.~\ref{fig:15}a) shows that the model
generalizes well when the testing resolution is equal to or finer than the training
resolution, consistent with \citet{bonatti2021}. The testing accuracy also generally improves
as the training resolution is refined. The behavior changes, however, when the same strain
paths are evaluated at coarser resolutions than those encountered during training: the error
can increase by several orders of magnitude. For example, a model trained on 500-increment
paths achieves an MSE of $6 \times 10^{-6}$ on the matching 500-increment test set but
degrades to $2 \times 10^{-2}$ on the corresponding 125-increment representation. Thus, for
the MSC, path discretization is not merely a data-sampling choice; it can materially alter
the learned constitutive response.

The corresponding heatmap for the CSS model (Fig.~\ref{fig:15}b) shows a different pattern.
Across all combinations of training and testing resolutions, the CSS model maintains low MSE
values of order $10^{-6}$. This remarkable resolution invariance is partly attributable to
the zero-order hold discretization embedded in the CSS model's recurrent core, which yields
the exact solution to the governing ODE for a constant forcing term (here, the normalized
strain increment direction $\bm{n}$). Once the CSS model has learned a locally linear
response at fine resolution, it reproduces that linearity accurately at coarser increments as
well, a capability the MSC's approximate self-consistency mechanism does not guarantee.

This test directly targeted a numerical property of a learned constitutive operator rather
than only its interpolation accuracy on a fixed dataset. The same underlying piecewise-linear
loading paths are represented at different increment sizes, and the CSS response remains
essentially unchanged over the investigated range. The result therefore supports the use of
increment-scaled state-space evolution as a means of reducing discretization dependence in
data-driven constitutive updates.

\subsection{Parallel training efficiency}\label{sec:4.4}

The heart of the CSS model is the stack of S5 blocks. Each S5 block features a linear
recurrence (Eq.~\ref{eq:15}). Given the input sequence
$\{\tilde{\bm{h}}_i^n\} = [\tilde{\bm{h}}_i^1,\tilde{\bm{h}}_i^2,...,\tilde{\bm{h}}_i^N]$
and the corresponding matrices
$\{\tilde{\mathbf{A}}_i^n\} =
[\tilde{\mathbf{A}}_i^1,\tilde{\mathbf{A}}_i^2,...,\tilde{\mathbf{A}}_i^N]$ and
$\{\tilde{\mathbf{B}}_i^n\} =
[\tilde{\mathbf{B}}_i^1,\tilde{\mathbf{B}}_i^2,...,\tilde{\mathbf{B}}_i^N]$, the
recurrence reads
\begin{equation}\label{eq:32}
\bm{\chi}_i^n =
\begin{cases}
0 & n = 1 \\
\tilde{\mathbf{A}}_i^n\, \bm{\chi}_i^{n-1} + \tilde{\mathbf{B}}_i^n
\tilde{\bm{h}}_i^n & 1 < n \leq N
\end{cases}
\end{equation}
which can be evaluated efficiently on multiple processors in parallel using an
all-prefix-scans operation, providing a runtime scaling of $\mathcal{O}(\log(n_p))$, with
$n_p$ denoting the number of processors \citep{blelloch1990}. In contrast, the MSC model
features a non-linear recurrence. Even though Eq.~\eqref{eq:11} is of seemingly linear
nature, it defines a non-linear recurrence as the update gate $\bm{g}$ also depends on the
previous state. The MSC recurrence thus needs to be evaluated in a sequential manner, i.e.,
it is not directly parallelizable, resulting in a runtime scaling of $\mathcal{O}(n_p)$.
This parallel-processing advantage of the CSS at the recurrence level is of interest when
the entire input sequence is known. This is not the case when the model is deployed within
a step-by-step finite element computation, but it is highly advantageous during training,
where the full strain-increment sequences (and hence also the $\{\tilde{\bm{h}}_i^n\}$
sequences) are known \textit{a priori}.

\begin{figure}[H]
  \centering
  \includegraphics[width=0.52\linewidth]{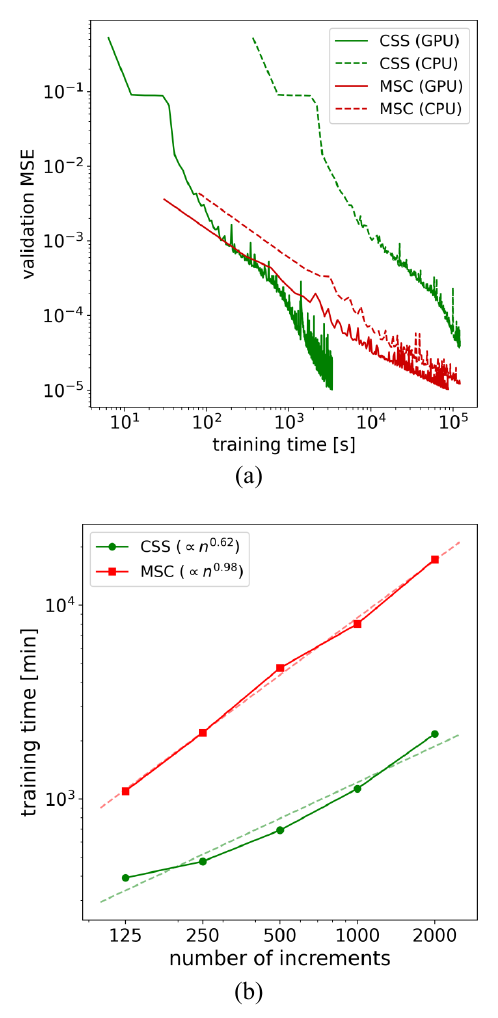}
  \caption{Scaling of training time: (a) validation loss vs time for training on four CPUs (dashed curves) and an RTX 4090 GPU (solid lines) for MSC and CSS models featuring roughly the same number of parameters, using the same dataset and the same batch size. (b) Time scaling for training on datasets of different sequence lengths (from 125 to 2,000 increments).}
  \label{fig:16}
\end{figure}

To compare training times, the MSC and CSS models are both trained on the mixed hardening
dataset with 125 increments and 8,000 training samples, on eight AMD EPYC 7742/9654 CPUs
(double-precision floating-point performance) as well as on an RTX 4090 GPU
(single-precision floating-point performance). The training used a batch size of 64 and
targeted a loss of $10^{-5}$. The MSC starts training at a lower loss value, but it
decreases much more slowly than that of the CSS (Fig.~\ref{fig:16}a). Both architectures
train faster on the GPU, with loss evolution scaling with training time independent of the
hardware employed. On the one hand, the CSS greatly benefits from GPU training, with the
training time reduced by two orders of magnitude for the same loss value; on the other
hand, the MSC's training time is only roughly halved when moving from the CPU to GPU
(Fig.~\ref{fig:16}a).

Focusing on the scaling of the training time with sequence length, and plotting the
training time for datasets with 125, 250, 500, 1000, and 2000-increment sequences
(Fig.~\ref{fig:16}b), the MSC's training time scales approximately with $n^{0.98}$, while
the CSS scales as $n^{0.62}$ (where $n$ is the sequence length). In other words, the CSS
can handle long sequences more efficiently than the MSC: for 125-increment sequences, the
CSS trains roughly $2.8\times$ faster than the MSC; at 2,000 increments, the gap widens to
roughly $8\times$, with the MSC requiring nearly 12 days of training versus under 1.5 days
for the CSS on the same dataset and hardware.

The parallelization benefit concerns sequence processing during training, when the complete
strain history is available and the linear recurrence can be evaluated by an associative
scan. In a step-by-step finite element analysis, future increments are not available and
the constitutive update is evaluated sequentially for both architectures. The computational
contribution is therefore a substantial reduction of the offline cost of learning
constitutive operators from long histories, rather than a claim of parallel-in-time
acceleration of the subsequent finite element solve.

\subsection{Training-data efficiency}\label{sec:4.5}

\begin{figure}[H]
  \centering
  \includegraphics[width=\linewidth]{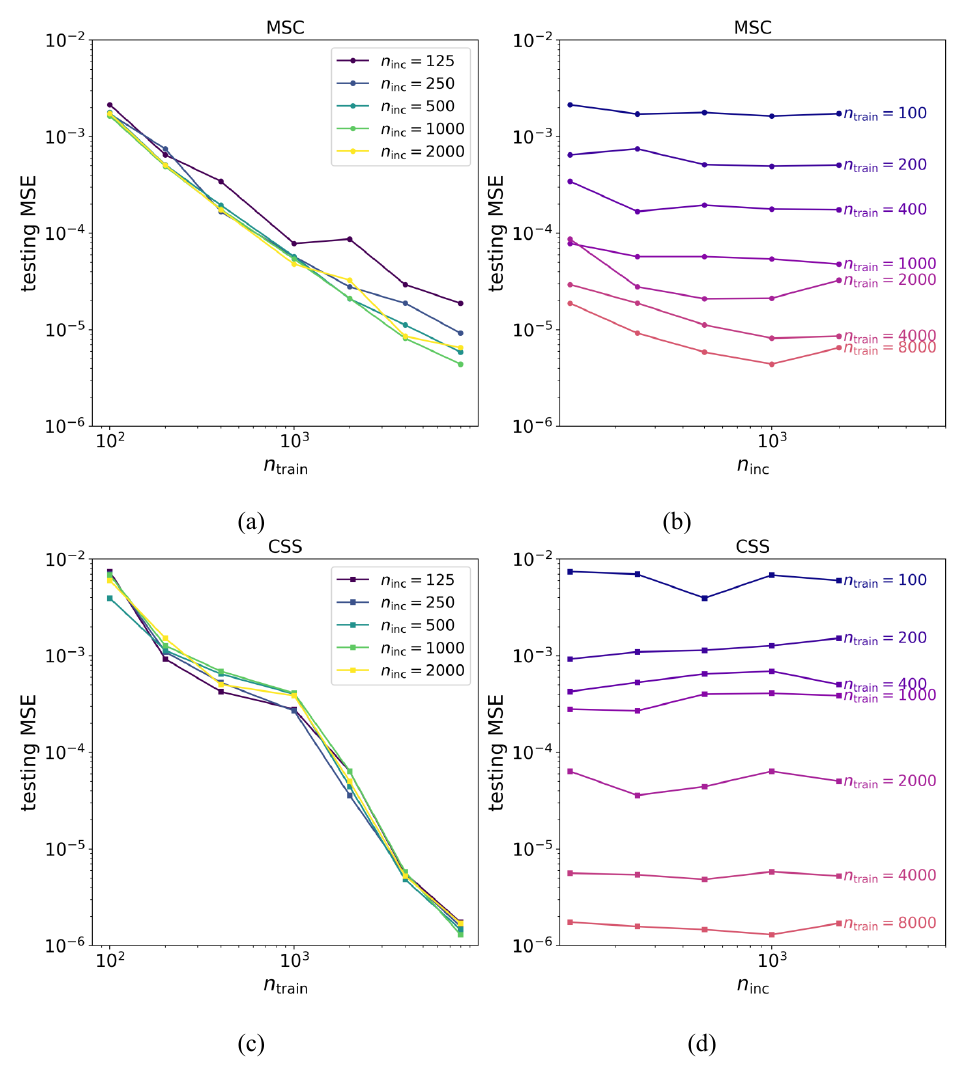}
  \caption{Effect of training dataset. The left plots (a, c) show the testing MSE as a function of the number of sequences included in the training dataset, while the right plots (b, d) show the length of the sequences (number of increments) on the abscissa axis. The top row plots (a, b) depict the results from the MSC model, while the bottom row results (c, d) are obtained with the CSS model.}
  \label{fig:17}
\end{figure}

The MSC and CSS testing errors are also examined as functions of the number of training
paths and the sequence length. The number of training paths is varied over
$\{100, 200, 400, 1000, 2000, 4000, 8000\}$ at sequence lengths of 125, 250, 500, 1,000,
and 2,000 increments. The testing sets are held fixed at 1,000 paths, while matching the
sequence length of the training sets. The MSC's testing error decreases monotonically with
the number of training paths $n_{train}$ (Fig.~\ref{fig:17}a), but not necessarily with the
sequence length $n_{inc}$ (Fig.~\ref{fig:17}b). The model achieves its lowest testing MSE
after training at intermediate resolutions (500--1,000 increments), rather than at the
longest sequences. For 8,000 training paths, the MSE is $4.39 \times 10^{-6}$ at 1,000
increments and increases to $5.86 \times 10^{-6}$ at 500 and $6.50 \times 10^{-6}$ at 2,000
increments, while the coarsest resolution (125 increments) yields $1.88 \times 10^{-5}$.
The gap widens for smaller datasets: when training on 2,000 paths only, the 125-increment
resolution gives an MSE of $8.69 \times 10^{-5}$ compared to $2.12 \times 10^{-5}$ at 1,000
increments, indicating that coarser temporal discretization requires substantially more
training data to compensate for the information lost between increments.

The CSS model's testing error also decreases monotonically with the number of training
paths across all resolutions (Fig.~\ref{fig:17}c), scaling favorably up to 8,000 paths. For
8,000 training sequence pairs, the CSS achieves an MSE of $1.30 \times 10^{-6}$ at 1,000
increments and remains between $1.47 \times 10^{-6}$ and $1.74 \times 10^{-6}$ at all other
resolutions tested. Overall, the CSS models are markedly more data-efficient: at 8,000
paths and the coarsest resolution (125 increments), the CSS reaches
$1.74 \times 10^{-6}$ MSE from only 1 million strain--stress pairs (8,000 sequences, 125
increments each), surpassing the MSC's best $4.39 \times 10^{-6}$ obtained from 8 million
pairs (8,000 paths at 1,000 increments), that is an $8\times$ reduction in total training
data at improved accuracy.

\subsection{Physical interpretability}\label{sec:4.6}

\begin{figure}[H]
  \centering
  \includegraphics[width=\linewidth]{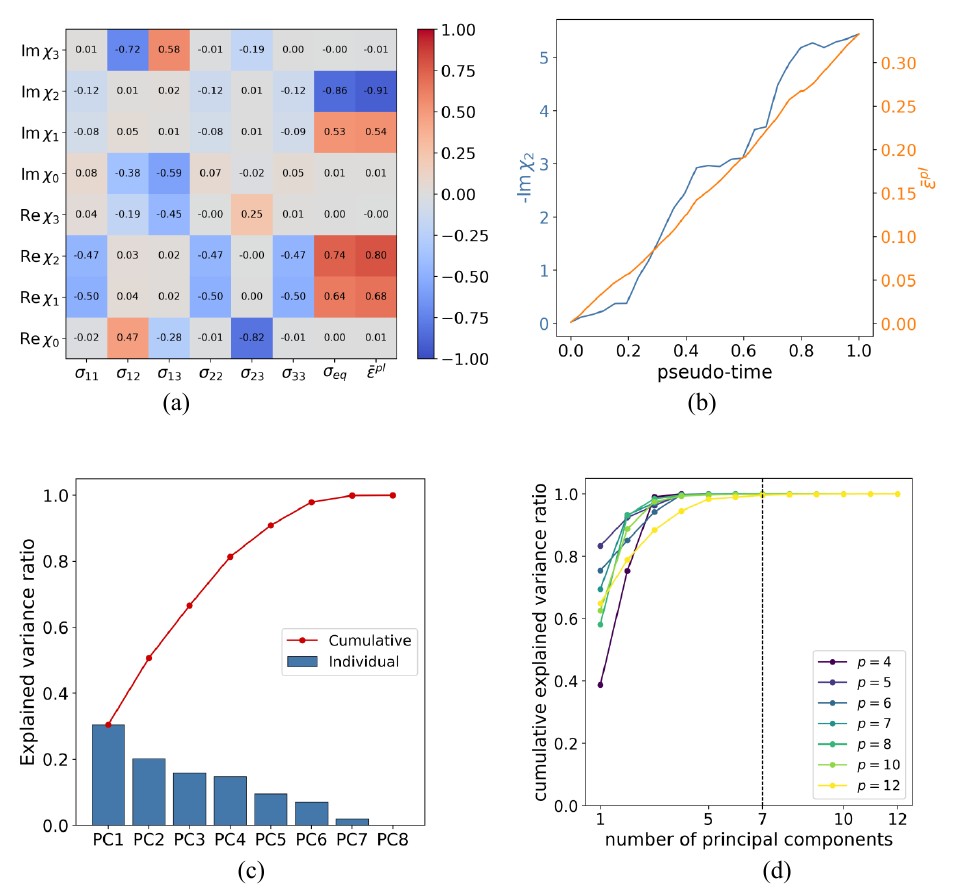}
  \caption{Interpretability of the CSS model's state variables: (a) Pearson's correlation coefficient (mean over the 1000 paths on the testing dataset) for a model with $d = 1$ and $p = 4$ (i.e. $s = 8$ real-valued state variables); (b) example of correlation of state variable $\mathrm{Im}(\chi_2)$ with the equivalent strain for a single path. (c) principal component analysis (PCA) on state space for CSS model with $d = 1$ and $p = 4$. (d) repeated PCA on models with larger complex state spaces (up to $s = 24$ real-valued state variables).}
  \label{fig:18}
\end{figure}

The interpretability of the MSC's state variables has been demonstrated by
\citet{bonatti2021}. To probe whether the learned internal representations of the CSS model
encode physically meaningful information, a single-layer CSS model is trained on the copper
dataset, with a complex state dimension $p = 4$, i.e., a total of $s = 8$ real-valued state
variables, and a latent input dimension of 107. The heatmap in Fig.~\ref{fig:18}a depicts
the Pearson coefficients $r$ that characterize the linearity of the relationship between
the CSS model's state variables (real and imaginary parts of $\bm{\chi}_0, ..,\bm{\chi}_3$)
and those of the training data generating Mises plasticity model (the stress-tensor
components and the equivalent plastic strain $\bar{\varepsilon}^{pl}$, plus the equivalent
stress $\sigma_{eq}$). The equivalent stress and equivalent plastic strain are most
strongly correlated with $\mathrm{Im}\,\chi_2$ ($r = -0.86$ and $-0.91$) and
$\mathrm{Re}\,\chi_2$ ($r = 0.74$ and $0.80$), followed by $\mathrm{Re}\,\chi_1$ and
$\mathrm{Im}\,\chi_1$ ($r = 0.53 - 0.68$) (Fig.~\ref{fig:18}a). Among the shear components,
$\mathrm{Re}\,\chi_0$ strongly correlates with $\sigma_{23}$ ($r = -0.82$),
$\mathrm{Im}\,\chi_3$ with $\sigma_{12}$ ($r = -0.72$), and $\mathrm{Im}\,\chi_0$ and
$\mathrm{Im}\,\chi_3$ with $\sigma_{13}$ ($r = -0.59$ and $0.58$). By contrast, no single
state variable correlates strongly with the normal stress components. Examining the
equivalent plastic strain $\bar{\varepsilon}^{pl}$ further, the best correlated CSS state
variable, $-\mathrm{Im}\,\chi_2$, tracks $\bar{\varepsilon}^{pl}$ with good fidelity across
the entire loading history for the paths corresponding to the 95\% (high-error) best
correlation (Fig.~\ref{fig:18}b). Principal component analysis (PCA) of the CSS models'
state trajectories reveals that the CSS model representing the copper material operates on
a seven-dimensional subspace, matching the number of state variables of the underlying
data-generating physics-based constitutive model: the variance-ratio plot of the principal
components (Fig.~\ref{fig:18}c) shows that only the first seven components carry
appreciable variance. Repeating this PCA for trained CSS models with larger complex state
space dimensions $p \in [5, 6, 7, 8, 10, 12]$ leads to the same conclusion
(Fig.~\ref{fig:18}d).

%% file: sections/05-conclusions.tex
\section{Conclusions}\label{sec:5}

This work introduced the Constitutive State Space (CSS) model, a mechanics-tailored
structured state-space formulation for data-driven modeling of path-dependent plasticity.
Its key computational feature is the decomposition of the physical strain increment into
magnitude and direction, with the increment magnitude entering directly into the
zero-order-hold discretization of a continuous-time linear state-space recurrence. This
construction yields a constitutive update that is stationary under zero increments, highly
robust to changes in strain-path resolution, and amenable to parallel prefix-scan
evaluation during training. The CSS therefore goes beyond a direct application of the S5
architecture to stress--strain sequences by adapting the state-space evolution to the
incremental structure of constitutive mechanics.

Across four benchmark materials including isotropic $J_2$-plasticity, pressure-sensitive
foam plasticity, and combined isotropic--kinematic hardening, the CSS matched or exceeded
the prediction accuracy of the MSC. For the plastically incompressible materials, the
validation loss was reduced by up to approximately one order of magnitude. More importantly
from a computational-mechanics perspective, the CSS maintained consistently low test errors
across all investigated combinations of training and testing strain-path resolutions,
whereas the MSC error increased by several orders of magnitude when evaluated at coarser
resolutions than those encountered during training. This reduced sensitivity to the
strain-increment discretization is particularly relevant for subsequent deployment in
incremental numerical simulations. The CSS also exhibited more favorable computational
scaling with sequence length, training up to eight times faster than the MSC for histories
containing up to 2,000 increments, while requiring substantially fewer strain--stress pairs
to achieve comparable or better prediction accuracy.

Finally, correlation and principal component analyses revealed physically meaningful
structure in the learned CSS states: individual state components correlate strongly with
quantities such as the equivalent plastic strain, while the effective dimensionality of the
learned state space is consistent with that of the underlying physics-based constitutive
model. Taken together, these results highlight three distinct requirements for learned
constitutive models that are often considered only jointly: approximation accuracy at a
fixed discretization, robustness of the constitutive update with respect to changes in
increment size, and computational efficiency when learning from long loading histories. The
CSS provides improvements in all three aspects without prescribing phenomenological
internal variables or their evolution equations.

Mechanics-tailored structured state-space models therefore provide a promising
computational alternative to nonlinear recurrent architectures for learned constitutive
modeling, particularly for applications involving long loading histories, variable
strain-increment sizes, or limited training data. Several extensions remain to be explored.
Incorporating strain-rate and temperature dependence would broaden the applicability of the
CSS to manufacturing and impact problems, while transfer-learning and multi-task learning
strategies may further reduce the amount of data required to adapt the formulation to new
materials. An important next step is its deployment in finite-element boundary-value
problems, where the consequences of the observed resolution robustness can be assessed
under heterogeneous deformation fields and adaptive load stepping. Extending the framework
to surrogate models of representative volume elements, including crystal-plasticity and
porous-plasticity homogenization, would further test its scalability and its ability to
represent material systems characterized by larger sets of physically meaningful internal
variables.

%% file: sections/06-backmatter.tex
\section*{Acknowledgements}

Valuable discussions with Dr. Xueyang Li (inspire/ETH Zurich) are gratefully
acknowledged.

\section*{CRediT authorship contribution statement}

\textbf{Rui Barreira:} Conceptualization, Data curation, Formal analysis,
Investigation, Methodology, Software, Visualization, Validation, Writing --
original draft, Writing -- review \& editing. \textbf{Taylan Soydan:}
Investigation, Formal analysis, Methodology, Software, Validation, Writing --
original draft. \textbf{Francesco Scipione:} Software, Investigation.
\textbf{Miguel Bessa:} Conceptualization, Resources, Methodology.
\textbf{Dirk Mohr:} Funding acquisition, Project administration, Supervision,
Visualization, Writing -- review \& editing.

\section*{Declaration of generative AI in the manuscript preparation process}

The authors used ChatGPT 5.5 for selected rephrasing. The author reviewed and
edited the output as needed and take full responsibility for the content of
the published article.